\documentclass{article}

\usepackage{microtype}
\usepackage{graphicx}
\usepackage{subfigure}
\usepackage{booktabs}
\usepackage{diagbox}
\usepackage[inkscapeformat=pdf]{svg}

\usepackage{hyperref}

\usepackage[accepted]{icml2023}

\usepackage{amsmath}
\usepackage{amssymb}
\usepackage{mathtools}
\usepackage{amsthm}
\usepackage{pdfpages}
\usepackage{relsize}

\usepackage[capitalize,noabbrev]{cleveref}

\theoremstyle{plain}

\theoremstyle{definition}

\theoremstyle{remark}

\usepackage[textsize=tiny]{todonotes}

\icmltitlerunning{Can Forward Gradient Match Backpropagation?}

\begin{document}

\twocolumn[
\icmltitle{Can Forward Gradient Match Backpropagation?}



\icmlsetsymbol{equal}{*}

\begin{icmlauthorlist}
\icmlauthor{Louis Fournier}{equal,yyy}
\icmlauthor{Stéphane Rivaud}{equal,yyy}
\icmlauthor{Eugene Belilovsky}{comp}
\icmlauthor{Michael Eickenberg}{sch}
\icmlauthor{Edouard Oyallon}{yyy}
\end{icmlauthorlist}

\icmlaffiliation{yyy}{Sorbonne Université, CNRS, ISIR, Paris, France}
\icmlaffiliation{comp}{CCM, Flatiron Institute, New York, USA}
\icmlaffiliation{sch}{MILA, Concordia University, Montréal, Canada}

\icmlcorrespondingauthor{Louis Fournier}{louis.fournier@isir.upmc.fr}
\icmlcorrespondingauthor{Stéphane Rivaud}{stephane.rivaud@isir.upmc.fr}

\icmlkeywords{Machine Learning, ICML, Local Learning, Decoupled Learning, Forward Gradient}

\vskip 0.3in
]

\printAffiliationsAndNotice{\icmlEqualContribution}

\begin{abstract}
Forward Gradients - the idea of using directional derivatives in forward differentiation mode  - have recently been shown to be utilizable for neural network training while avoiding problems generally associated with backpropagation gradient computation, such as locking and memorization requirements.
The cost is the requirement to guess the step direction, which is hard in high dimensions.
While current solutions rely on weighted averages over isotropic guess vector distributions,
we propose to strongly bias our gradient guesses in directions that are much more promising, such as feedback obtained from small, local auxiliary networks.
For a standard computer vision neural network, we conduct a rigorous study systematically covering a variety of combinations of gradient targets and gradient guesses, including those previously presented in the literature.
We find that using gradients obtained from a local loss as a candidate direction drastically improves on random noise in Forward Gradient methods.

\end{abstract}

\section{Introduction}
Stochastic Gradient Descent (SGD)~\cite{sgdamari4039068, bottou2012stochastic} using end-to-end backpropagation for gradient computation is the ubiquitous training method of state-of-the-art Deep Neural Networks. However, one can identify at least two issues with this gradient estimation procedure. For one, it requires a significant amount of computational resources and memory. Indeed, a backpropagation update is performed in two steps: a forward step which computes activations, and a backward step which computes gradients. Until a gradient at a given layer is computed, intermediary computations which lead to this layer must be stored in memory, and computation of the update is blocked. This is referred to as the \emph{backward lock} \cite{jaderberg2016decoupled}, which is also an obstacle in the development of model-parallel asynchronous procedures \cite{belilovsky2021decoupled}. Backpropagation also lacks biological plausibility, which is often considered a direction towards more scalable learning \cite{ren2022scaling}. Indeed, 
the backpropagation algorithm suffers from the weight transport problem~\cite{carpenter1989neural}, which means that the weights of the above layers must be shared during the backward pass of a given layer. This requires a significant amount of communication and synchronization, which is not observed to be implementable in biological systems~\cite{liao2016important}. Since biological systems have achieved learning without these restrictions, researchers hope to isolate and use such principles to improve artificial neural network learning.

Recent contributions have re-examined the converse procedure of computing \emph{Forward Gradients} \cite{baydin2022gradients}, built on the classical forward mode automatic differentiation \cite{baydin2018automatic}.
It is an alternative to standard backpropagation algorithm, which allows the computation of a directional derivative (i.e., a scalar product with the gradient, as opposed to the gradient vector itself) from a forward pass.
The gradient computations in forward mode explicitly use the Jacobian of a given layer through Jacobian-Vector Products but do not require storing any intermediary activations or  backward passes. In other words, it is \emph{backward unlocked} (as described by \citet{jaderberg2016decoupled}) in that the computation of the derivative is finished as soon as the forward pass is completed. This leads to a potential memory reduction and does not use  the biologically implausible weight transport. A Forward Gradient, as introduced by \citet{baydin2022gradients}, corresponds to an unbiased estimate of an activation or weight gradient (which we will refer to as a Gradient Target) computed via a random, isotropic, directional derivative, i.e., a projection of the Gradient Target onto a direction vector (which we will define as the Gradient Guess). The motivation of Forward Gradient descent is to approximate End-to-End Gradient descent. However, this unbiased gradient approximation generally suffers from high variance.

Several approaches \cite{ren2022scaling,silver2021learning} have been proposed to reduce the corresponding variance.  Unfortunately, there is often an accuracy gap~\cite{silver2021learning} which is difficult to reduce because the Gradient Guess is not aligned with the End-to-End Gradient Target. Following a different line of ideas, \citet{ren2022scaling} propose to approximate the gradients computed from Local Loss functions \cite{pathak2022local,nokland2019training,belilovsky2019greedy} as Gradient Targets. This is combined with a collection of architectural changes allowing the decomposition of losses into subgroups within a layer~\cite{patel2023local}. This reduces dimensionality and, thereby, the variance of Forward Gradients. However, these changes lead to architectures that are not broadly applicable. Furthermore, results are still far from competitive with standard backpropagation. 

Following \citet{ren2022scaling}, it is also unclear if using local gradients as Gradient Targets combined with a Gaussian Gradient Guess is the optimal strategy. 
In this work, we propose to study
different combinations of Gradient Targets and Gradient Guesses in a way that covers existing work and also explores other possible combinations. More specifically, we investigate whether local gradients, which alone lead to suboptimal performance, can be used to compute a reliable Forward Gradient estimate of the end-to-end gradient.
Our goal is to understand better if the forward mode of automatic differentiation can lead to training competitive models while maintaining the above benefits. We believe our study to be of high interest due to our in-depth coverage of possible update signal sources on a well-established neural network architecture.

\begin{figure}[ht]
          \centering
          \includegraphics[width=.7\linewidth]{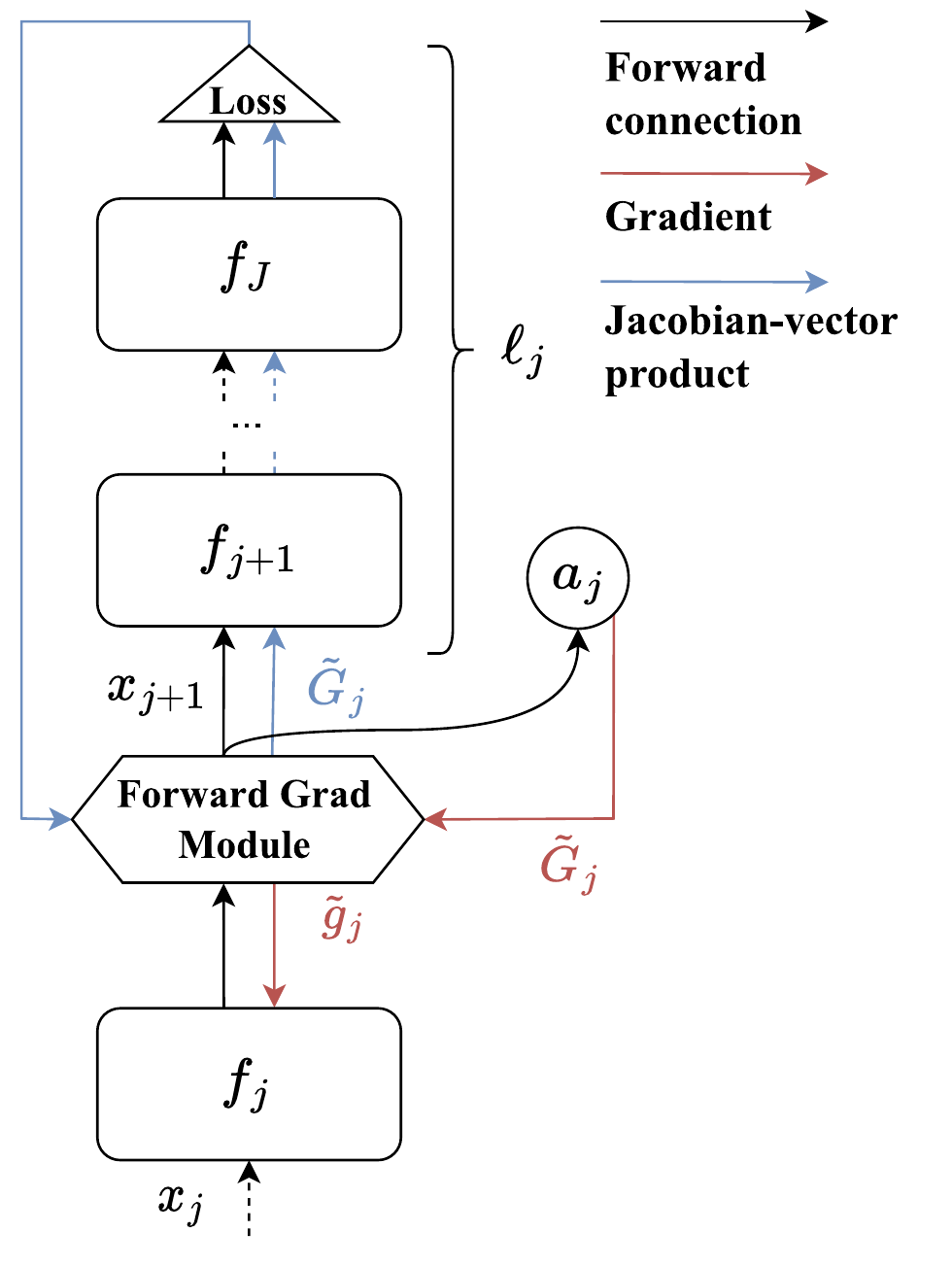}\label{fig:schema}
          \caption{Schematic summarizing the best use-case of our algorithm, which approximates a Global (End-to-End) Target gradient at a block $j$ with forward mode automatic differentiation. A  Gradient Guess  $\tilde G_j$ is provided by the gradient of local loss $a_j$ (red). The Gradient Target is projected on the Gradient Guess by Jacobian-vector products (blue), and we use this estimate to compute the update for  block $j$. Also, in \citet{ren2022scaling}, the Guess $\tilde G_j$ is a random vector. More details can be found Sec. \ref{sec:gradient-approx}}
        \end{figure}

\paragraph{Contributions.} Our contributions are as follows:  \textbf{(a)} 
We make explicit the idea that any Gradient Target can be projected onto any Gradient Guess and propose to study a wide variety of these combinations.
\textbf{(b)} In particular, we propose to use local gradients from auxiliary losses as powerful Gradient Guess directions for the computation of directional derivatives in Forward Gradient mode.
We show that this proposal can yield far improved results compared to naive random directions while maintaining the benefits of the Forward Gradient mode. \textbf{(c)} To ablate this method, we cover many combinations of possible  Gradient Targets and Guesses for a well-established architecture (ResNet-18) applied to standard image classification, using both gradient insertion points commonly described as \textit{activity perturbation} and \textit{weight perturbation}, which indicate whether the directional derivative is taken in activation space or weight space.
 \textbf{(d)} Our evaluations reveal that in the case of Gradient Guesses obtained from supervised
Local Losses,
a consistent positive alignment between the Gradient Targets and Guesses improves the Forward Gradient estimation, reducing the gap with end-to-end training.

Our paper is structured as follows: Sec. \ref{sec:gradient-approx} describes Gradient Target approximations with Forward Gradient, using both Random and Local Gradient Guesses. Then, Sec. \ref{sec:experiments-setup} describes all the details of our experimental setup. In Sec. \ref{sec:insight1}, we show that Forward Gradient can lead to competitive accuracy with end-to-end training, and in Sec. \ref{sec:insight2} that this is due to having a better estimate of the Global Target than with Random Guesses. However, we show in Sec. \ref{sec:insight3} that the Local Gradient Guess is still a poor estimate due to different training dynamics between Local and Global losses. Finally, we confirm the consistency of our findings across model sizes and other datasets in Sec. \ref{consistency}. Our source code is available at: \href{https://github.com/streethagore/ForwardLocalGradient}{github.com/streethagore/ForwardLocalGradient}.

\section{Related work}

\paragraph{Forward Gradient via forward mode automatic differentiation.} Forward Gradient has been recently popularized by \citet{baydin2022gradients}, which showed that using a random direction for the input of forward mode automatic differentiation gives an unbiased estimate of the gradient. They 
studied Forward Gradient descent on example datasets and obtained promising results. Our work attempts to study and understand the limits of Forward Gradient descent on a well-established architecture (Resnet-18) and a difficult computer-vision task (ImageNet32). Note that further theoretical analysis of Forward Gradient was done by \citet{belouze2022forwardgrad}, where they show that a Rademacher random direction offers the best unbiased estimate, but again with no deep learning experiments. Forward Gradient has also been studied in the context of RNNs~\cite{silver2021learning}, as it is also an ideal candidate to solve the issue with long-term Backpropagation-through-time. This technique has been applied to small benchmarks, but the question of its benefits on challenging tasks remains open.

\paragraph{Forward Gradient for biologically plausible models.} Since Forward Gradient allows the computation of a directional derivative in forward mode, this routine does not require any weight transport or waiting for an update signal as in backpropagation and can be implemented via one Forward pass. Forward Gradient is thus a training mechanism that is more biologically plausible than the Backpropagation mechanism. 
To scale Forward Gradient, \citet{ren2022scaling} proposes three techniques that improve the approximation properties of Forward Gradient: architecture modifications that introduce more separability and decorrelation (similar to \citet{patel2023local}), local losses, and to apply Forward Gradient on activations rather than on parameters for reducing the variance of the gradient estimate.

\paragraph{Local losses.} In essence, Forward Gradient allows the removal of the backward lock, as introduced by \citet{jaderberg2016decoupled}, that is inherent to the backpropagation algorithm. Before updating its weights, a given block need not wait for a full backward pass. This idea has been extensively tested in \citet{belilovsky2021decoupled,shallow,lowe2019,nokland2019training,gomez2022interlocking} by using local losses, which also remove other computational locks, increasing the potential for asynchrony of the training procedure. However, such methods fail to reach the same performance as the end-to-end backpropagation when too much parallelism is desired: incorporating feedback, such as the feedback of the Forward Gradient, is a promising direction for reducing the gap in the accuracy of such methods. In our work, we will use gradients from local losses as candidates of directions for computing Forward Gradients.

\section{Gradient approximation via Forward Gradient}\label{sec:gradient-approx}

The objective of this paper is to study computationally efficient and accurate estimates $g(x)\in \mathbb{R}^d$ of a \textbf{Gradient Target} $\nabla f(x)\in \mathbb{R}^d$ of some  objective function $f:\mathbb{R}^d\to \mathbb{R}$, using Forward Gradients. The main idea is to use a \textbf{Gradient Guess} vector $G(x)\in \mathbb{R}^d$ onto which we project the Gradient Target to obtain an approximation:
\begin{align}
\label{eq:grad_estimate}
g(x) = \langle \nabla f(x), G(x) \rangle G(x)\approx \nabla f(x)\,.
\end{align}
Once $G(x)$ is provided, then $g(x)$ can be efficiently computed as it can be implemented as a Forward Gradient. This induces limited computational overhead compared to a standard forward pass since evaluating the directional derivative costs around as much as a forward pass (especially for non-linearities whose derivatives cost as much as the function itself). The choice of $G$ will thus greatly impact the quality of the approximation. Note that while Gradient Descent aims at determining $x^*\in \mathbb{R}^d$ such that $\nabla f(x^*)=0$, the stationary points of \eqref{eq:grad_estimate} are given by:
\begin{align*}G(x^*)=0\text{ or }\langle \nabla f(x^*),G(x^*)\rangle=0\,.\end{align*}

\subsection{Gradient Guesses}
Gradient Guesses come in different flavors of randomness. Hence
the term ``approximation" describing \eqref{eq:grad_estimate} can take on both probabilistic and deterministic forms. We discuss guesses that are purely random, purely deterministic, and one intermediate case.

\paragraph{Random Guesses.} 
By drawing $G$ from a zero-mean, unit-covariance probability distribution, we can create an unbiased estimator of the gradient.
We propose to use either Rademacher variables $\mathbb{P}(G_i=1)=\mathbb{P}(G_i=-1)=\frac 12$ as proposed by \citet{belouze2022forwardgrad}, or isotropic Gaussians $G\sim \mathcal{N}(0,\mathbf{I})$, as proposed in \citet{ren2022scaling}. In both cases, $g$ is an unbiased estimate of $\nabla f$ as we have $\mathbb{E}[GG^T]=\mathbf{I}$. However, such estimates potentially have high variance, leading to large errors in individual gradient estimates and slow optimization progress.

\paragraph{Local Guesses.}
To improve the quality of this estimate, we consider deterministic gradient guesses. We obtain these guesses by computing local update signals from local auxiliary losses, as in \citet{belilovsky2021decoupled,nokland2019training}. This gives access to a gradient $\nabla a(x)$ from a small local model $a(x)$ with minimal computational effort. 
 Multiple choices for $a(x)$ are possible, for example, a CNN, an MLP, or a linear layer. The intuition behind using local gradients is that they should provide a better one-shot approximation of $\nabla f(x)$ than uncorrelated noise. To obtain the best linear approximation of $\nabla f(x)$ from $\nabla a(x)$, we will then  use  $G(x)=\frac{\nabla a(x)}{\Vert \nabla a(x)\Vert}$ as the projection direction.

 \paragraph{NTK Guesses.}
Conceptually, comparing random and deterministic projection directions via local auxiliary losses is not straightforward. We include a transitional setting using random auxiliary networks to test whether the inductive bias of the random network is sufficient for improvement or whether training of the auxiliary is required.
 To do so, we re-initialize the Local Model $a(x)$ introduced above at each iteration of the algorithm. This corresponds to a Guess obtained from the Neural Tangent Kernel (in finite width) of our model~\cite{jacot2018neural}. Note that this can be understood as a weaker version of DFA~\cite{nokland2016direct,bartunov2018assessing}, which we expand on in App. \ref{fixntk}.

\subsection{Gradient Targets}
\paragraph{Global Target.} To update any set of weights, the most common gradient target in a supervised Neural Network setting is the gradient of the loss of the final layer of our model.
The use of this gradient is standard practice~\cite{krizhevsky2009learning}, and it is present in most Deep Learning models. 
For Forward Gradients, \citet{baydin2018automatic} uses this gradient as the target, as illustrated in Fig. \ref{fig:schema}. We will study this setting extensively, and also call this Target the End-to-End Gradient Target.

\paragraph{Local Target.}
However, it is also possible to take the gradient from the auxiliary loss of the current layer as a gradient target, in order to perform fully local learning.
\citet{ren2022scaling} uses such Local Targets in their experiments and obtains their best results with it. We refer to local learning as the case where the Gradient Guess for the Local Target is equal to the Local Guess. (As for end-to-end training, this is a particular case of Forward Gradient with the Guess exactly equal to the Target).

\paragraph{Intermediate Target.}
One could also obtain
gradient targets from auxiliary losses attached to any intermediate layer of the model. 
Such a target is likely to be better correlated with the Local Guess while also being closer to the global target, possibly offering a middle ground.

\subsection{Gradient computation and insertion points}
The candidate direction for the directional derivative obtained using the Forward Gradient technique can be computed in different spaces, with two natural candidates: parameters or activations.

To propose an update of the $j$-th block $f_j(x_j,\theta_j)$ where $\theta_j$ describes the parameters of the $j$-th block and $x_j$ is the output of the block $f_{j-1}$, the strategy is to train a model via gradient descent. The output $f_j(x_j,\theta_j)$ is fed to a (target) loss $\ell_j$,
where $\ell_j$ can represent a local loss obtained by attaching an auxiliary network to the current or a subsequent block, or it can represent the standard loss at the output of the network. The procedure uses a sample estimate $\nabla_{\theta_j} (\ell_j\circ f_j)$ of the gradient aggregated into
a mini-batch $\mathcal{B}=\{x^1_j,...,x^n_{j}\}$ of data:
\begin{align}
\nabla_{\mathcal{B}}(\ell_j\circ f_j)(\theta_j)&\triangleq \frac{1}n\sum_{i=1}^n\nabla_{\theta_j}(\ell_j\circ f_j)(x^i_j,\theta_j)\label{minibatch}\\
&=\frac{1}n\sum_{i=1}^n\partial_{\theta_j} f_j(x^i_j,\theta_j)^T\nabla_{x_{j+1}}\ell_j(x^i_{j+1})\label{minibatchacti}
\end{align}
\paragraph{Weight perturbation.}
A first natural strategy is to use the so-called weight perturbation of the gradient, which means that the estimate obtained in Eq. \eqref{minibatch} is replaced by:
\begin{equation}g_j(x)=\frac{1}n\langle \nabla_{\mathcal{B}}(\ell_j\circ f_j)(\theta_j),\frac{1}n\sum_{i=1}^nG^i_j\rangle \sum_{i=1}^nG^i_j\,
\end{equation}
where $G^i_j$ is a gradient guess for $\nabla_{\theta_j} (\ell_j\circ f_j)(x^i_j,\theta_j)$. 

\paragraph{Activity perturbation.}
However, if $G^i_j$ is pure noise, \citet{ren2022scaling} observed that this estimate has high variance. In this case, estimating the gradient via \textit{activity perturbation} is preferable. Indeed, observe that the update of Eq. \eqref{minibatchacti} can be approximated by:
\begin{equation}
\tilde g_j(x)=\frac{1}n\sum_{i=1}^n\partial_{\theta_j} f_j(x^i_j,\theta_j)^T\langle \tilde G^i_j,\nabla_{x_{j+1}}\ell_j(x^i_{j+1})\rangle \tilde G^i_j\label{eq-activity}
\end{equation}
where $\tilde G^i_j$ is a  guess for $\nabla_{x_{j+1}}\ell_j(x_{j+1}^i)$, and which is often of dimension smaller than $G^i_j$, leading to a reduced variance in the case of e.g., pure noise~\cite{ren2022scaling}.

\paragraph{Compute trade-off.}

We summarize in Table \ref{tab:computetradeoff} the unlocking capabilities as well as compute tradeoffs for the different pairs of Guesses and Targets we introduced. As described by \citet{jaderberg2016decoupled}, standard backpropagation is backward-locked and requires a global vector-Jacobian product. Forward Gradient however allows backward unlocking with a Global Target and even update unlocking with a Local Target. As noted in \citet{ren2022scaling}, weight perturbation has favorable memory usage, as activity perturbation requires storing intermediate activations for the Forward Gradient estimator. Activity perturbation also necessitates a backpropagation step in the associated block to compute the gradient used to update the weights.  

\begin{table}[!ht]
    \centering
    \caption{Unlocking and compute tradeoffs for different Guess and Target pairs. Here, vJp and Jvp stand for vector-Jacobian product and Jacobian-vector product respectively. The flag (local) means that the output of the operation (vJp or Jvp) for a given module is not needed by subsequent modules. }
    \begin{tabular}{|l|l|l|l|}
    \hline
     \multicolumn{2}{|l|}{Guess}  & Global Target & Local Target \\ \hline
        Local & Unlocking &  Backward & Update \\ \cline{2-4}
       \& NTK  & Compute & vJp (local)+Jvp & vJp (local) \\ \hline
        Random & Unlocking &  Backward & Update \\ \cline{2-4}
      & Compute  & Jvp & Jvp (local) \\  \hline
    \end{tabular}
    \label{tab:computetradeoff}
\end{table}

\section{Numerical experiments}
\subsection{Experimental setup}\label{sec:experiments-setup}
We now describe how we implemented our models, training procedure, and the implementations of Gradient Targets and Guesses to study the accuracy of a given model under the variations of those parameters. 
 \paragraph{Models. }For this set of experiments, we consider two models which allow us to test various hypotheses on the scalability of gradient computations in forward mode: a standard ResNet-18 \cite{he2016deep} (divided into 8 blocks with local losses) and the same ResNet-18 without skip-connections, to avoid side-effects when combining this model with layer-wise local losses (divided into 16 blocks with local losses to reach a fully layerwise subdivision of the network, except for the first block which contains two layers). 

 \paragraph{Hyper-parameters. } We considered the CIFAR-10 and ImageNet32 datasets, used with standard data augmentation. The ImageNet32 dataset is smaller than the full-resolution ImageNet dataset, making it more efficient to use during training and testing while still providing a challenging task for model performance. \citet{chrabaszcz2017downsampled} has demonstrated that, in general, the conclusions drawn from the ImageNet32 dataset are also applicable to the full-resolution ImageNet dataset. We followed a standard training procedure: SGD with a momentum of 0.9 and weight decay of $5\times 10^{-4}$.
 For CIFAR-10, we train the model for 100 epochs, with a learning rate decayed by $0.2$ every 30 epochs. For ImageNet32, we also first try a shorter training of 70 epochs, decaying the learning rate by $0.1$ every 20 epochs.
 The initial learning rate was chosen among $\{0.05, 0.01, 0.005\}$ for CIFAR-10 and $\{0.1, 0.05, 0.01, 0.005, 0.0001\}$ for ImageNet32.
 We report the validation accuracy at the last epoch of training using the best training procedure and learning rate. More details are given in App. \ref{appendix}.

 \begin{figure}[ht]
\label{fig:train-test-loss2}
          \centering
          \includegraphics[width=.99\linewidth]{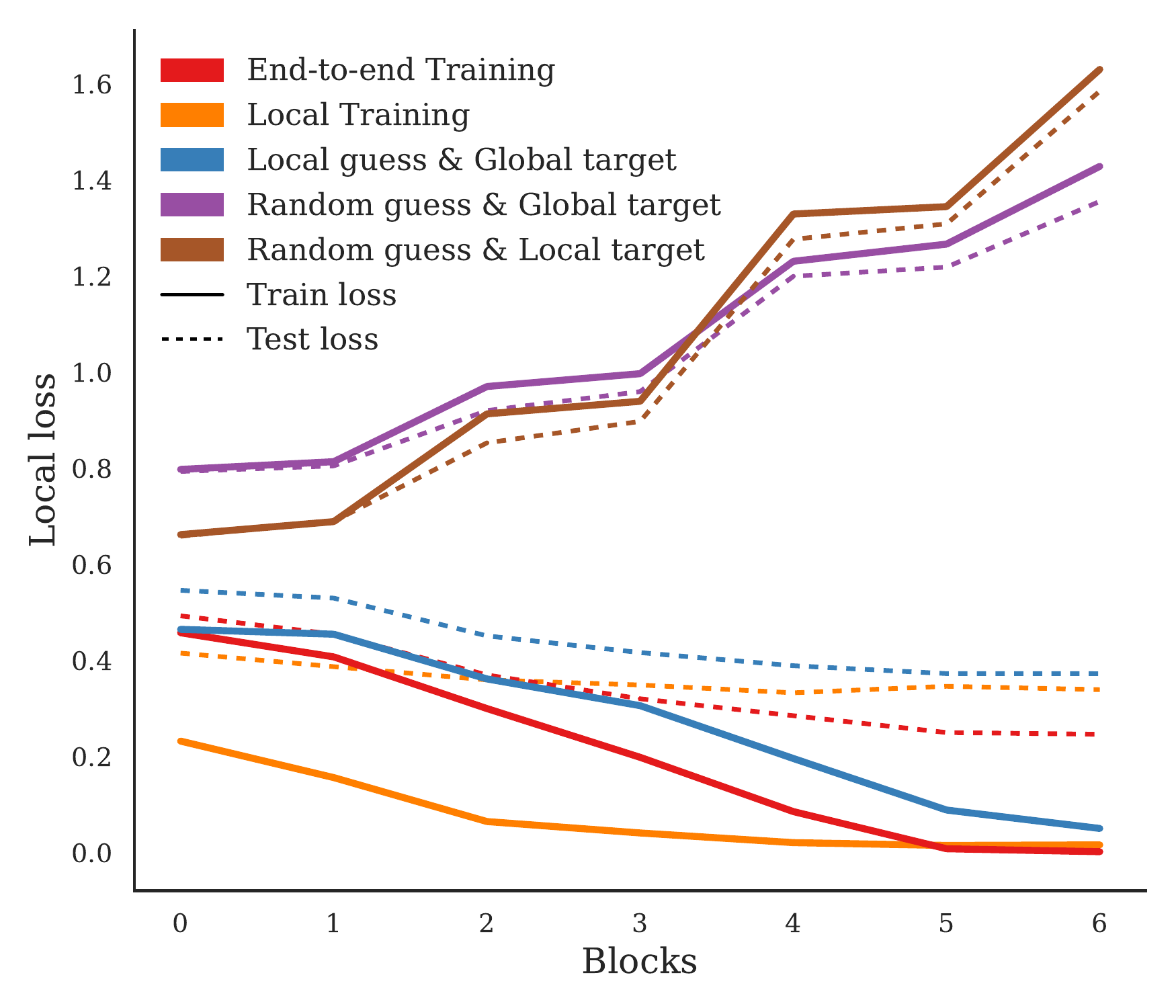}
          \caption{Local train losses at the end of training at each block for a ResNet-18 split in 8 blocks, with CNN auxiliary, for different training algorithms. In the Gaussian and end-to-end cases, the auxiliary training is detached from the main module training and is only for logging purposes. With Gaussian Guesses, losses increase with depth. The Local Target allows better convergence than the Global Target at the first block but an even worse convergence for the whole network. With local learning, local losses converge to local minima.
          }\label{fig:loss-progressive}
\end{figure}

\begin{table}[!ht]
\small
    \centering
    \caption{Test accuracy of a ResNet-18 using a Global Target, split into 16 local-loss blocks, on CIFAR-10 and ImageNet32 for both  activity (Activ.) and weight perturbations. Weight perturbation systematically outperforms activity perturbation with a Local Guess. The MLP auxiliary provides the best Local Guess in all configurations. We report the mean and standard deviation over 4 runs for CIFAR-10.}
\label{tab:resnet18-cifar10-imagenet32}

\resizebox{\columnwidth}{!}{\begin{tabular}{|l|c|c|c|c|}
\hline Dataset &  \multicolumn{2}{c|}{CIFAR-10} &  \multicolumn{2}{c|}{ImageNet32}\\ \hline
End-to-End &\multicolumn{2}{c|}{94.3 ± 0.1}  &\multicolumn{2}{c|}{53.7}\\\hline
Space & Activ.  & Weight & Activ. & Weight \\ \hline
 Local, CNN                          & 79.0 ± 1.0 & 88.0 ± 0.4 & 7.3 & 40.0 \\
 Local, MLP                          & 84.7 ± 0.3 & 88.7 ± 1.2 & 21.8 & 37.4  \\ 
  Local, Linear                      & 46.7 ± 2.5 & 86.1 ± 1.4 & 10.0 &  23.3  \\\hline 
NTK, CNN                    & 37.7 ± 0.1 & 50.1 ± 1.0 & 2.0 & 3.6  \\ 
 NTK, MLP                    & 50.3 ± 0.4  &  49.7 ± 0.6  & 7.5 & 3.9   \\
 NTK, Linear                      & 49.9 ± 1.0    & 48.3 ± 0.7 & 4.2 & 3.9   \\\hline
 Gaussian    & 38.9 ± 0.9 & 50.0 ± 0.8 & 4.9 & 4.9 \\
 Rademacher  & 38.0 ± 1.5 & 49.8 ± 0.2 & 5.5 & 4.6  \\ \hline 
\end{tabular}}

\end{table}

 \paragraph{Local models and losses.} We considered 3 types of trainable local auxiliary models: a CNN, an MLP, and a Linear Layer, designed and developed for use with CIFAR-10, to add no more than $10\%$ compute overhead (in FLOPS) while leading to good accuracy. Again, we used a subset of the training data to cross-validate this architecture, which we then fixed during our experiments.  Our CNN local auxiliary model is a 3-layer CNN with ReLU non-linearities followed by a $2\times 2$ adaptive average pooling and a projection onto the classification space. The MLP is a 3-layer MLP using ReLU non-linearities, followed by a projection on the classification space. The linear auxiliary net consists of a $2\times 2$ adaptive average pooling preceded by a batch normalization module and followed by a projection onto the classification space as in \citet{belilovsky2021decoupled}. More details on our methodology to cross-validate the hyper-parameters can be found in App. \ref{appendix}. Each local loss can then be used to obtain Gradient Guesses via an abridged backpropagation procedure or to serve as Gradient Target for computing Forward Gradients.

\subsection{The gap between backpropagation and Forward Gradient is narrowed with Local Guesses
}\label{sec:insight1}

Tab. \ref{tab:resnet18-cifar10-imagenet32} and Tab. \ref{tab:resnet18-block-cifar10} report our results for a ResNet-18 split into 8 or 16 blocks, on both CIFAR-10 and ImageNet32. First, we note that accuracies of Forward Gradient with a Random Guess as studied in \citet{ren2022scaling} are extremely low.  Despite extensive hyper-parameter searches (for learning rate notably), it proved difficult to achieve reasonable accuracies for ImageNet32 using random Gradient Guesses. This is in line with expectations from \citet{belouze2022forwardgrad},
and suggests that the architecture modifications performed in \citet{ren2022scaling} are essential to its success.

\begin{table}[!ht]
\small

    \centering
    \caption{Test accuracy of a ResNet-18 using a Global Target, split into 8 local-loss blocks, on CIFAR-10 and ImageNet32 for both activation (Activ.) and weight perturbations. Local Guesses improve on Random Guesses and NTK Guesses. Activity perturbation only provides an advantage for Random Guesses, otherwise, weight perturbation performs better in all CIFAR-10 configurations. We report the mean and standard deviation over 4 runs for CIFAR-10.}\label{tab:resnet18-block-cifar10}
\begin{tabular}{|l|c|c|c|c|}
\hline Dataset &  \multicolumn{2}{c|}{CIFAR-10} &  \multicolumn{2}{c|}{ImageNet32}\\ \hline
End-to-End & \multicolumn{2}{c|}{94.5 ± 0.2}& \multicolumn{2}{c|}{54.6}\\\hline
Guess & Activ. & Weight & Activ. & Weight \\ \hline
Local, CNN      & 90.4 ± 0.2 & 92.0 ± 0.3 & 33.1 & 38.3 \\ 
Local, MLP      & 89.2 ± 0.2 & 91.6 ± 0.2 & 38.0 & 45.8 \\
Local, Linear   & 65.9 ± 6.7 & 88.9 ± 0.3 & 27.9 &  34.9\\  \hline
NTK, CNN        & 55.2 ± 0.4 & 66.8 ± 2.6 &  5.5  & 8.0 \\
NTK, MLP        & 61.6 ± 0.5 & 63.1 ± 1.1 &  17.4 & 15.5 \\ 
NTK, Linear     & 61.6 ± 0.9   & 62.6 ± 1.0 & 9.8 & 15.4  \\
\hline
\end{tabular}
\end{table}

\begin{figure}[ht]
          \centering
          \includegraphics[width=.9\linewidth]{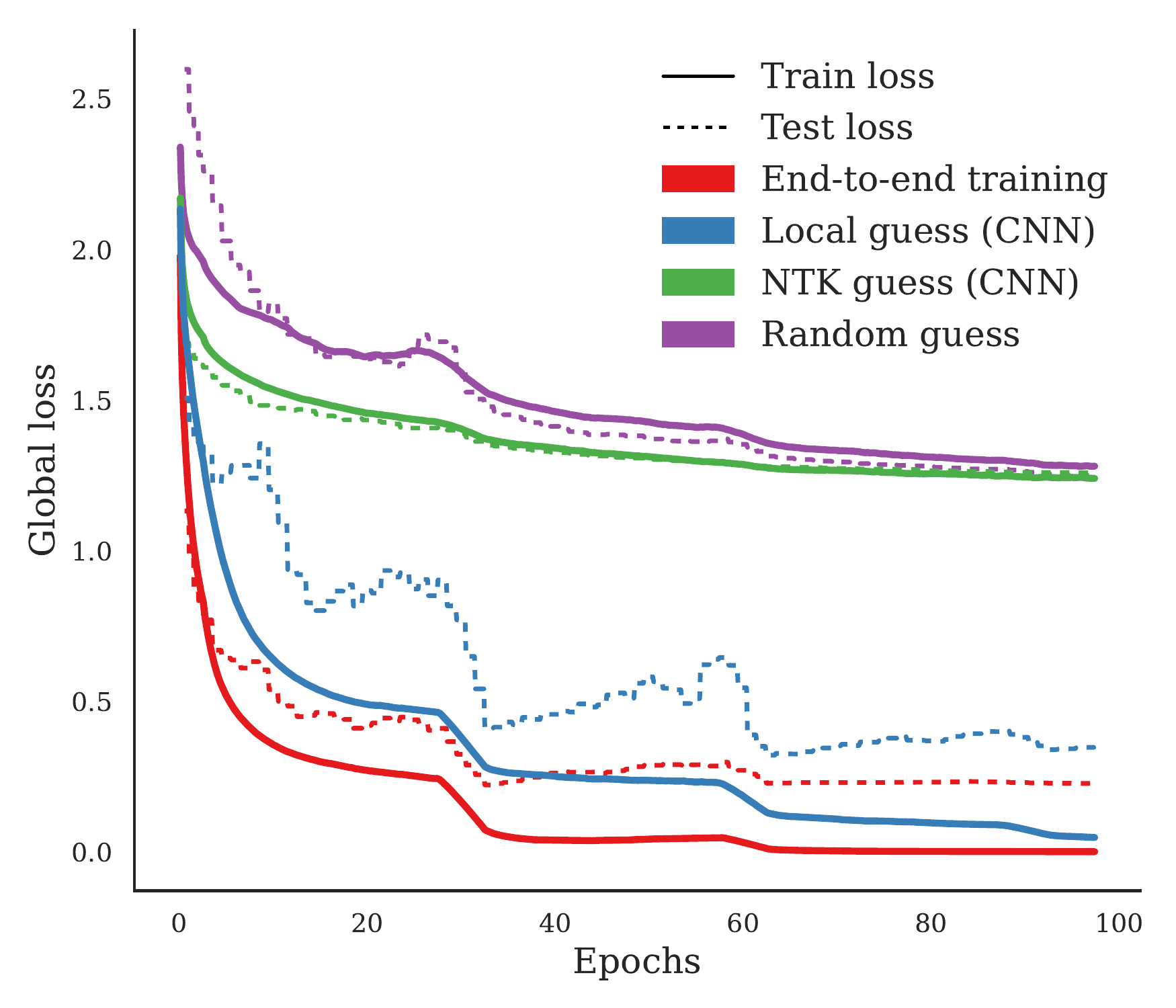}
          \caption{Comparison of train and test losses for end-to-end training (red), and Forward Gradient with Global Target and Local Guess (blue), NTK Guess (green) and Random Guess (purple), on CIFAR-10 with a ResNet-18 split in 8 blocks. Local and NTK Guesses are derived from a CNN auxiliary. Random and NTK Guesses fail to optimize the Global Loss. Local Guess performs much better, with a consistent gap with respect to end-to-end throughout training.}
\label{fig:train-test-loss}
\end{figure}

 \begin{figure*}[ht]
          \centering
          \includegraphics[width=.8\linewidth]{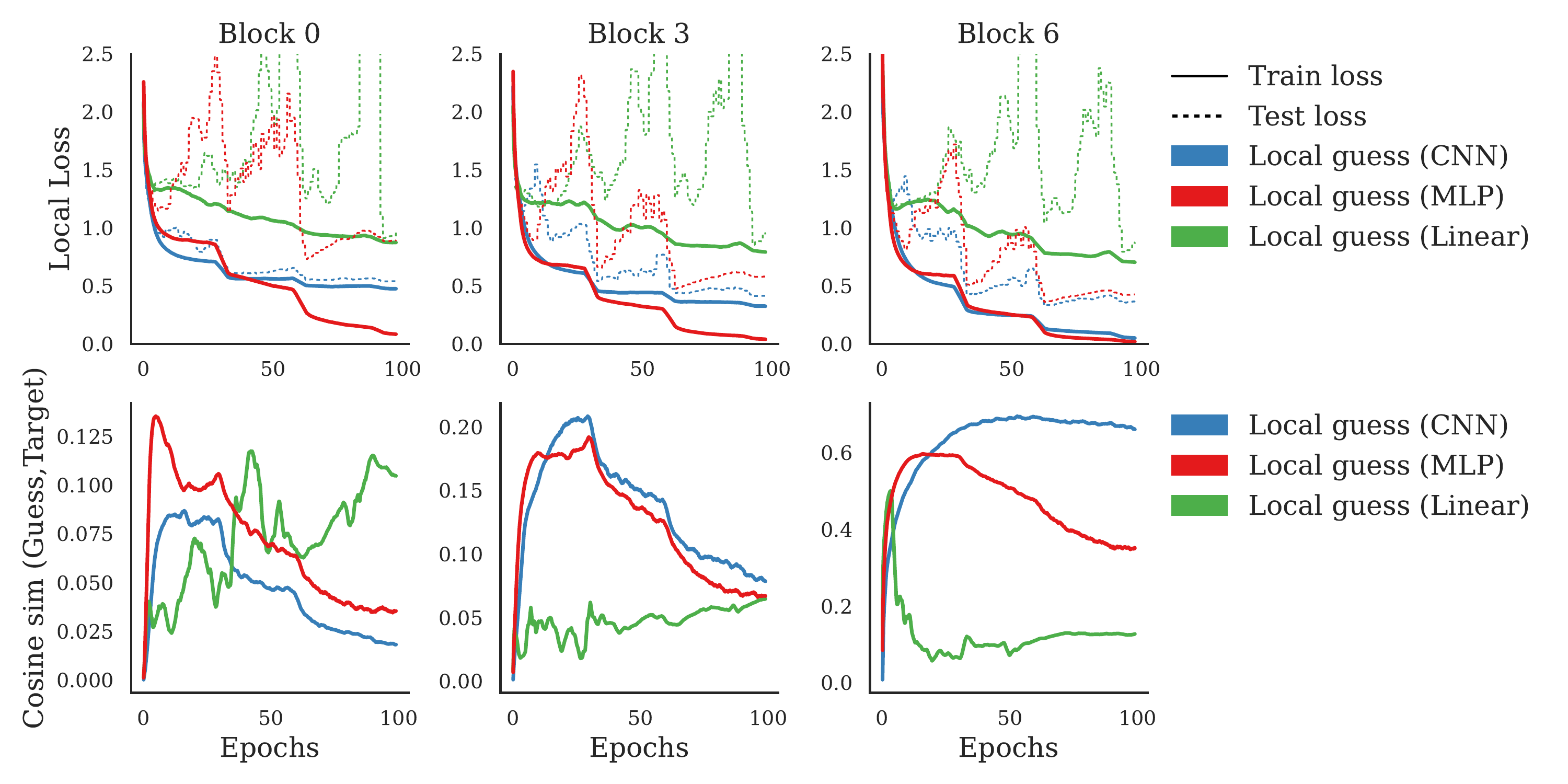}
          
          \caption{Train and test local losses (top row) and mean cosine similarity between Local Guess and Global Target in the activation space (bottom row), for blocks 0, 3, and 6 (left, middle, and right columns) during training. The model is a ResNet-18 divided into 8 blocks trained on CIFAR-10. The similarity values are low but consistently positive and increase with depth as the target gets closer (and less deep). Learning improves with depth, as expected.  As both target and local losses converge to a minimum, their similarity decreases.}\label{fig cossim activ}
        \end{figure*}

        \begin{figure*}[ht]
          \centering
          \includegraphics[width=.8\linewidth]{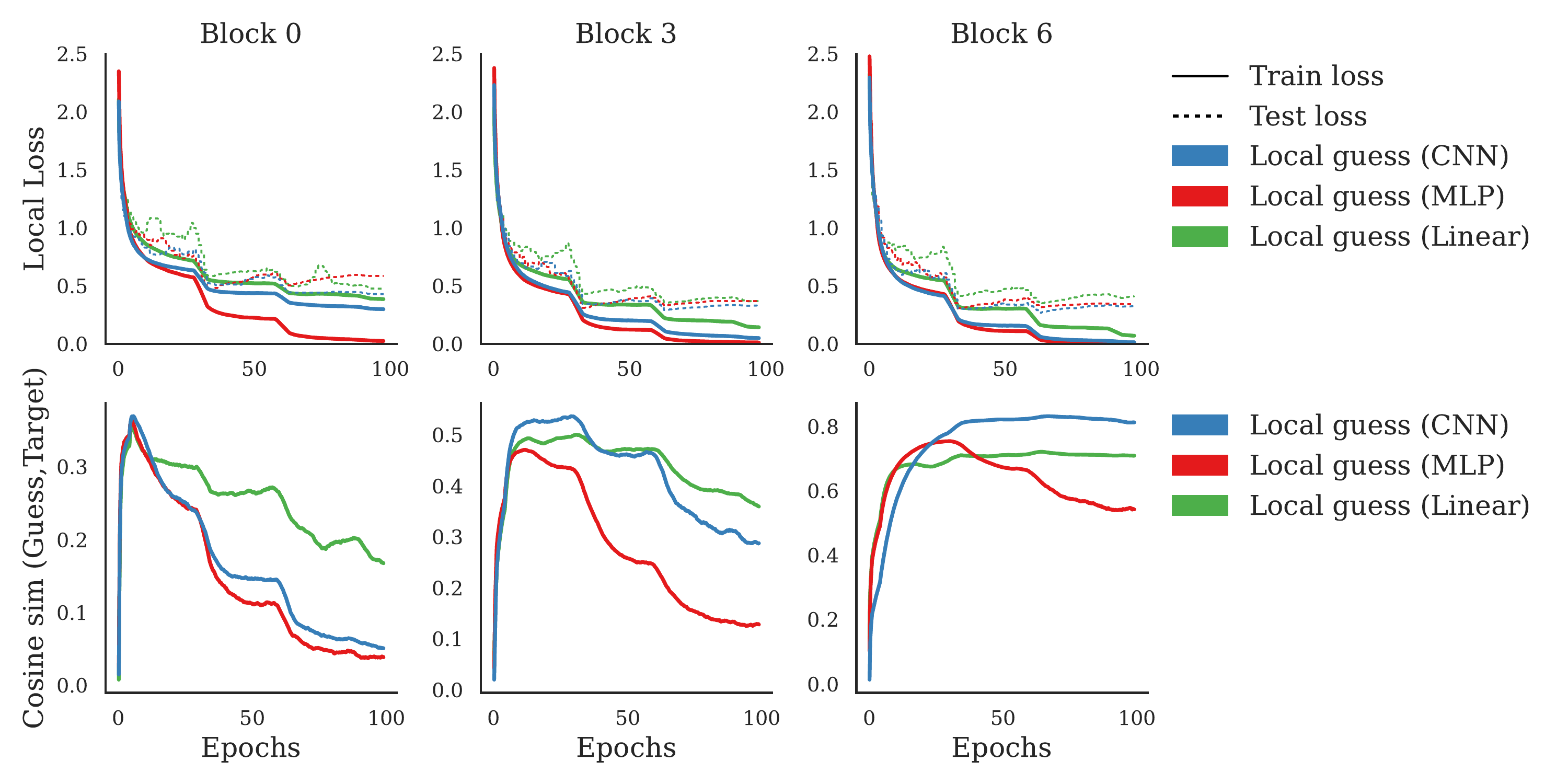}
          \caption{Train and test local losses (top row) and mean cosine similarity between Local Guess and Global Target in the weight space (bottom row, averaged over the parameters), for blocks 0, 3, and 6 (left, middle, and right columns) during training. The generalization gap is more consistent during training than with activation gradients. The cosine similarity is also consistently higher than for activation gradients but falls more drastically during training.}
          \label{fig cossim weight}
        \end{figure*}

\paragraph{Local Guesses reduce the gap with end-to-end training.}
Tab. \ref{tab:resnet18-cifar10-imagenet32} and Tab. \ref{tab:resnet18-block-cifar10} 
indicate that Local Guesses systematically outperform the naive Random Guesses strategy, in some cases by about $40\%$ for CIFAR-10 and $20\%$ for ImageNet32. 
This is not surprising, as a pure Gaussian Guess leads to random directions which are not aligned with the directions important for the classification task.
Even NTK Guesses, which are random but benefit from the inductive bias of the auxiliary network, yield a slight improvement in most cases.

Reusing the same random network for a full mini-batch permits us to obtain a gradient direction which is less naive than random noise and proves that NTK ~\cite{jacot2018neural} has a good inductive bias for this task. 
We report in Figure \ref{fig:train-test-loss} the train and test losses for Forward Gradient training with Global Target and several Guesses, showing that only Local Guesses allow convergence of the global loss.

\paragraph{Weight perturbation outperforms activity perturbation. 
}
For the case of supervised gradient guesses, we note that weight perturbation significantly outperforms activity perturbation. This shows that the best setup for the setting studied in \citet{ren2022scaling} may not generalize to standard neural networks.

\begin{table}[!ht]
    \centering
    \caption{Test accuracy on CIFAR-10 of each variation of a ResNet-18, using a Local Target with Gaussian, Rademacher, and Local Guesses, for both activity and weight perturbations. We report the mean and standard deviation over 4 runs.}
    \label{tablocal}
\resizebox{\columnwidth}{!}{\begin{tabular}{|l|c|c|c|c|c|c|}
\hline
8 blocks       & \multicolumn{2}{c|}{ CNN} & \multicolumn{2}{c|}{MLP} & \multicolumn{2}{c|}{Linear} \\\hline
Local Target        & \multicolumn{2}{c|}{92.0 ± 0.2} & \multicolumn{2}{c|}{92.1 ± 0.2} & \multicolumn{2}{c|}{89.2 ± 0.1} \\  \hline
Guesses               & Activity & Weight  & Activity & Weight   & Activity & Weight  \\ \hline
Gaussian            & 41.2 ± 1.9 & 52.1 ± 1.5 & 45.6 ± 1.6 & 53.2 ± 0.6 & 36.9 ± 1.9 & 56.5 ± 0.8 \\
Rademacher          & 39.0 ± 1.3 & 53.0 ± 0.7 & 43.6 ± 2.1 & 53.3 ± 0.9 & 34.6 ± 1.9 & 56.6 ± 0.6 \\ \hline  \hline
16 blocks       & \multicolumn{2}{c|}{ CNN} & \multicolumn{2}{c|}{MLP} & \multicolumn{2}{c|}{Linear} \\\hline
Local Target
        & \multicolumn{2}{c|}{87.7 ± 0.1} & \multicolumn{2}{c|}{89.4 ± 0.3} & \multicolumn{2}{c|}{86.7 ± 0.3} \\  \hline 
Guesses             & Activity & Weight  & Activity & Weight   & Activity & Weight \\ \hline
Gaussian            & 23.5 ± 2.4 & 37.5 ± 1.6 & 21.5 ± 4.4 & 39.7 ± 0.8 & 23.8 ± 1.6 & 43.5 ± 0.5 \\
Rademacher          & 22.3 ± 1.5 & 37.8 ± 0.5 & 23.3 ± 1.4 & 41.3 ± 1.4 & 24.1 ± 1.3 & 45.5 ± 0.5 \\ \hline 
\end{tabular}}
\end{table}

\paragraph{Training by residual blocks obtains a significantly better accuracy than training layerwise} Comparing the results of Tab. \ref{tab:resnet18-cifar10-imagenet32} and Tab. \ref{tab:resnet18-block-cifar10} indicates that systematically, for the same pair of algorithms, training by blocks of two layers (corresponding to residual blocks) with one auxiliary per block outperforms the corresponding version with one auxiliary per layer.
This observation is consistent with the fact that the layers encapsulated in a given block are jointly trained in the first case.
By contrast, the layerwise strategy is fully decoupled. The fact that using  joint training systematically improves the accuracy of  the method is known from previous work \cite{belilovsky2021decoupled,patel2023local,belilovsky2019greedy}. However,
it comes at the cost of not obtaining the full potential of computational and memory savings associated with Forward Gradient.
For Forward Gradient methods, we note that Local Guesses with weight perturbation are the Guesses that minimize the accuracy drop when transitioning from a 8 blocks split to a 16 blocks one.

\subsection{Reliably estimating the Global Target is necessary for Forward Gradient to succeed}
\label{sec:insight2}
\paragraph{Accuracy degrades in  upper layers with Random Guesses.} Fig. \ref{fig:loss-progressive} shows the evolution of the separability of representations with depth via the training loss of a local (CNN) auxiliary for a ResNet-18 split into 8 blocks. In the case of end-to-end training and Random Guesses for Global targets, the auxiliary losses were trained ad-hoc without interacting with the training of the global model. Without any surprise, the end-to-end progressively separate classes, a phenomenon already observed in \citet{zeiler2014visualizing,jacobsen2018revnet,oyallon2017building}. Random isotropic Guesses do not follow this trend. On the contrary, the training loss progressively degrades with depth, both for Global and Local Targets.
 In comparison, Local Guesses with Global Target accuracies improve with depth, as we expect from a deep model. Furthermore, Random Guesses with Local Target give worse results than Random Guesses with Global Target, as reported in Tab. \ref{tablocal}. This constitutes a further difference between this work and \citet{ren2022scaling}.

\paragraph{Intermediate Targets perform worse than Global Targets.} To bridge the gap between Global and Local Targets, we proposed an Intermediate Target, the gradient of the Local loss at block $j+1$. The idea is to have a target closer to the Gradient Guess than the Global Target, allowing for easier convergence (since similarity increases with depth)
as well as improving on Local Target, where subsequent blocks do not provide any feedback. However, despite more consistent alignment between Targets and Guesses across blocks, we observed a consistent decrease in accuracy compared to both Global and Local Targets, as reported in Tab \ref{tab:resnet18-cifar10-j}. Thus, despite good alignment between Guess and Target, not using a Global Target penalizes the learning of the network.

\begin{table}[!ht]
\small
    \centering
    \caption{Test accuracy on CIFAR-10 of a Resnet-18 using the loss of the block $j+1$ as the Intermediate Target for the block $j$, for both  activity (Activ.) and weight perturbations. The model has been split into 8 and 16 blocks as in the experiments of Tab. \ref{tab:resnet18-block-cifar10} and Tab.  \ref{tab:resnet18-cifar10-imagenet32}\label{tab:resnet18-cifar10-j}. We added the Target (tgt) used for the Random Guesses.}
\resizebox{\columnwidth}{!}{\begin{tabular}{|l|c|c|c|c|}
\hline ResNet-18 split  &  \multicolumn{2}{c|}{8 blocks} &  \multicolumn{2}{c|}{16 blocks}\\ \hline
Gradient Guesses &Activ.  & Weight&Activ. &Weight \\ \hline
Local, CNN         & 91.6  & 91.6  & 80.0  & 87.8  \\ 
Local, MLP           & 91.5  & 91.8     & 87.6  & 89.5     \\
Local, Linear         & 76.3  & 89.6     & 59.8  & 86.9     \\ \hline 
NTK, CNN       & 55.7  & 60.0  & 33.0  & 18.4  \\
NTK, MLP     & 57.4  & 52.5  & 40.7 & 20.1  \\ 
NTK, Linear     & 55.3  & 44.2 &  32.4  & 20.3  \\\hline
Gaussian, CNN tgt                   & 45.3    & 51.5       & 28.0     & 47.1      \\
Gaussian, MLP tgt                & 53.7     &  53.2   & 31.4     &  48.0  \\
Gaussian, Linear tgt                 & 51.4    & 53.9      & 42.5    & 46.4     \\
Rademacher, CNN tgt             & 43.8    & 51.6     & 24.8    & 47.5     \\ 
Rademacher, MLP tgt              & 54.7     & 52.9     &  30.4   & 49.5     \\ 
Rademacher, Linear tgt                 & 49.0     & 54.2    & 43.6   & 45.6     \\  \hline
\end{tabular}}
\end{table}

\subsection{Local Guesses do not align with the Global Target} \label{sec:insight3}

We now study the difference between the optimization path of the Forward Gradient and the end-to-end gradient to understand if Forward Gradient can be understood as an approximation of the Global Target i.e., the End-to-End Gradient Target. 

\paragraph{Local Guesses and Global Targets are weakly aligned.}

Fig. \ref{fig cossim activ} and \ref{fig cossim weight} display the evolution of the local train and test losses, and the cosine similarity between Local Guesses and the Global Target, for a ResNet-18 split into 8 blocks. 
For the cosine similarity, we report a batch estimate at every iteration given by:
\begin{equation}
     \frac{1}{|\mathcal{B}|} \mathlarger{\sum}_{i \in \mathcal{B}} \frac{\langle \nabla_{x_{j+1}}\ell_j(x^i_{j+1}),\tilde G^i_j \rangle}{\Vert \nabla_{x_{j+1}}\ell_j(x^i_{j+1})\Vert \Vert \tilde G^i_j \Vert}\,.
\end{equation}

We note that the cosine similarity between the end-to-end gradient and the gradients obtained from local auxiliaries show three main tendencies: \textbf{(a)} With increasing block index, the alignment is higher (see y-axis scale on each plot).
\textbf{(b)} Most of the time, the convolutional auxiliary shows the strongest correlation and the lowest test loss value. \textbf{(c)} Finally, although low, the similarity is consistently positive.
The first statement is a reflection of the depth proximity of the auxiliary to the Global Target; the second is likely due to the convolutional nature of the subsequent network.
We further note that MLP and linear auxiliaries share similarly erratic test loss curves and that most similarity curves tend to descend toward the end of training. The former may be due to an architectural misalignment. 
The latter observation likely indicates that different local minima have been reached and that the optimization trajectories will, from then, drift apart.

\paragraph{Projecting on the entire Gradient Guess space still does not estimate the Gradient Target.}
The low alignment between Local Guesses and Global Targets indicates a poor approximation. We propose projecting each Global Target on the subspace spanned by Local Guesses, for each sample to test this hypothesis further. In other words, we replace the gradient estimate of Eq. \eqref{eq-activity} with the best linear approximation of the Global Target using (the entire subspace of) Local Guesses. In practice, for a batch of gradients, we project each Global Target onto the span of the batch of Local Guesses. 
Our results are aggregated in Tab. \ref{tab-span}, and we trained each model using this novel dynamic.
The improvement over projecting on a single Local Guess is marginal, and the accuracy is still far from end-to-end training. This means that approximating the Global Target by Local Guesses is insufficient to recover the dynamic of backpropagation training. Generally, the task of approximating a Global Target by Gradient Guesses is likely a difficult task \cite{montufar2014number}.

\begin{table}[!ht]
\small 
    \centering
    \caption{Test accuracy on CIFAR-10 of a Resnet-18 trained by using the best linear approximation of the end-to-end gradients by local guesses, with activity perturbation. We report the mean and standard deviation over 4 runs.}\label{tab-span}
\begin{tabular}{|l|l|c|} 
\hline
Model & Gradient Guess subspace &Accuracy  \\ \hline
ResNet-18,  & Local Guess, CNN   & 92.3 ± 0.1 \\
 8 blocks& Local Guess, MLP      & 89.6 ± 0.3   \\
& Local Guess, Linear            & 84.0 ± 0.6  \\ \hline 
ResNet-18,& Local Guess, CNN     & 89.6 ± 0.6   \\
16 blocks&Local Guess, MLP       & 77.9 ± 3.3  \\
& Local Guess, Linear            & 73.4 ± 1.8  \\ \hline 
\end{tabular}
\end{table}

\begin{table*}[!h]
\small
    \centering
    \caption{Test accuracy of a ResNet-18 using a Global Target, split into 8 or 16 local-loss blocks, on Fashion-MNIST, CIFAR-100 and Imagenette datasets for both activity and weight perturbations. We report the mean and standard deviation over 4 runs.}
\label{tab:imagenette-global}
\resizebox{\textwidth}{!}{
\begin{tabular}{|l|c|c|c|c|c|c|c|c|c|c|c|c|}
\hline Dataset &   \multicolumn{4}{c|}{CIFAR-100}  &  \multicolumn{4}{c|}{Fashion-MNIST} & \multicolumn{4}{c|}{Imagenette}     \\
\hline Model &  \multicolumn{2}{c|}{16 blocks} &  \multicolumn{2}{c|}{8 blocks}  &  \multicolumn{2}{c|}{16 blocks} &  \multicolumn{2}{c|}{8 blocks} &  \multicolumn{2}{c|}{16 blocks} &  \multicolumn{2}{c|}{8 blocks}  \\ \hline
End-to-End &\multicolumn{2}{c|}{74.6 ± 0.3} &\multicolumn{2}{c|}{75.7 ± 0.2}  &\multicolumn{2}{c|}{94.1 ± 0.2} &\multicolumn{2}{c|}{93.9 ± 0.1}  &\multicolumn{2}{c|}{88.9 ± 0.4} &\multicolumn{2}{c|}{90.0 ± 0.4} \\\hline
Guess & Activity & Weight & Activity & Weight  & Activity & Weight  & Activity & Weight  & Activity & Weight  & Activity & Weight  \\ \hline
 Local, CNN         & 44.6 ± 1.7 & 58.2 ± 0.6 & 64.5 ± 0.9 & 67.9 ± 0.4 & 92.5 ± 0.3 & 93.7 ± 0.1 & 93.7 ± 1.3 & 93.6 ± 0.2 & 71.8 ± 1.8 & 84.2 ± 0.6 & 84.9 ± 0.3 & 88.0 ± 0.1 \\
 Local, MLP         & 59.5 ± 0.5 & 64.3 ± 0.4 & 67.5 ± 0.3 & 70.0 ± 0.3 & 92.3 ± 0.2 & 93.8 ± 0.1 & 93.6 ± 0.3 & 94.0 ± 0.1 & 77.3 ± 1.3 & 82.8 ± 0.3 & 84.9 ± 0.4 & 85.9 ± 0.4\\ 
 Local, Linear      & 46.1 ± 1.1 & 62.1 ± 0.4 & 50.0 ± 9.8 & 65.9 ± 0.4 & 73.4 ± 1.7 & 93.4 ± 0.2 & 87.4 ± 1.5 & 94.0 ± 0.1 & 53.2 ± 2.0 & 82.8 ± 0.4 & 70.7 ± 1.7 & 86.1 ± 0.5 \\\hline 
 NTK, CNN           & 15.0 ± 0.7 & 22.9 ± 0.3 & 29.9 ± 0.7 & 40.3 ± 1.5 & 82.0 ± 0.2 & 88.2 ± 0.3 & 87.2 ± 0.4 & 91.1 ± 0.3 & 47.0 ± 1.0 & 53.7 ± 2.1 & 65.9 ± 0.7 & 68.2 ± 1.0 \\ 
 NTK, MLP           & 23.5 ± 0.3 & 23.1 ± 0.4 & 35.9 ± 1.0 & 36.1 ± 0.2 & 86.8 ± 0.6 & 88.0 ± 0.5 & 88.3 ± 0.3 & 90.8 ± 0.2 & 56.3 ± 0.8 & 53.3 ± 0.9 & 69.7 ± 1.0 & 67.9 ± 0.7 \\
 NTK, Linear        & 23.9 ± 0.5 & 22.9 ± 0.7 & 36.4 ± 0.4 & 37.8 ± 2.2 & 83.7 ± 0.2 & 88.8 ± 0.3 &  88.5 ± 0.4  & 90.8 ± 0.3 & 55.3 ± 1.4 & 51.3 ± 0.8 & 68.7 ± 0.6 & 68.6 ± 0.5 \\\hline
 Gaussian   & 10.0 ± 1.2 & 22.4 ± 0.5 & 27.5 ± 0.5 & 35.3 ± 0.5 & 79.1 ± 0.7 & 88.3 ± 0.3 & 87.5 ± 0.2 & 89.5 ± 0.3 & 17.8 ± 1.5 & 47.6 ± 1.4 & 52.7 ± 3.1 & 64.1 ± 1.1 \\
 Rademacher & 8.7 ± 1.1 & 22.6 ± 0.4 & 27.5 ± 0.4 & 33.2 ± 1.2 & 79.5 ± 0.2 & 88.3 ± 0.5 & 87.1 ± 0.7 & 89.7 ± 0.2 & 18.4 ± 2.6 & 47.9 ± 1.7 & 51.8 ± 1.7 & 63.8 ± 0.9 \\ \hline 
\end{tabular}
}
\end{table*}

\paragraph{Alignment between Guess and Target matters.} Experimental results indicate that the Local Guess does not approximate well the Global Target.
Comparing Tab. \ref{tablocal} and Tab. \ref{tab-span} shows that using the best linear approximation of the Global Target on the space spanned by the Local Guesses still performs on par or worse than local learning (i.e., Local Guess on Local Target).
Thus, despite the Global Target being a better target than the Local Target in general, the adequation between the Guess and the Target has a stronger impact than the reweighting of the Guess.
However, it is known that local learning saturates at early layers and does not benefit from the depth of the model \cite{wang2021revisiting}, indicating that global feedback may be necessary to recover backpropagation performances.
These issues suggest that deriving a good estimate of the end-to-end gradient is a difficult task that might need to be addressed to obtain performance on par with standard backpropagation.

\begin{table}[h]
\small
    \centering
    \caption{Test accuracy of a Wide ResNet-18 using a Global Target, split into 16 local-loss blocks, on CIFAR-10 for both activity and weight perturbations for different width factors, i.e., k=0.5, 2 and 4. Table \ref{tab:resnet18-cifar10-imagenet32} refers to the case where $k=1$. We report the mean and standard deviation over 4 runs.}
\label{tab:wideresnet-global}
\resizebox{1.\columnwidth}{!}{\begin{tabular}{|l|c|c|c|c|c|c|}
\hline 
Width factor k &   \multicolumn{2}{c|}{$0.5$}  &  \multicolumn{2}{c|}{$2$} & \multicolumn{2}{c|}{$4$}     \\ \hline
End-to-End &\multicolumn{2}{c|}{93.0 ± 0.4}  &\multicolumn{2}{c|}{94.9 ± 0.0} &\multicolumn{2}{c|}{95.3 $\pm$ 0.1}\\\hline
Guesses & Activity  & Weight & Activity & Weight & Activity  & Weight \\ \hline
 Local, CNN & 63.3 ± 1.6 & 72.4 ± 0.6 & 87.0 ± 0.4 & 85.5 ± 0.3 & 90.6 ± 0.2 & 93.3 ± 1.5 \\
 Local, MLP & 81.5 ± 0.5 &  79.1 ± 0.3 & 86.2 ± 0.2 & 84.1 ± 0.2 & 87.0 ± 0.2 & 91.7 ± 0.2 \\ 
 Local, Linear & 26.9 ± 3.0 & 78.5 ± 0.3 & 62.7 ± 2.8 & 84.6 ± 0.2 & 62.5 ± 4.3 & 92.3 ± 0.3 \\\hline 
 NTK, CNN      & 33.8 ± 1.3 & 48.9 ± 0.2 & 41.1 ± 0.4 & 50.6 ± 0.3 & 44.2 ± 0.3 & 51.6 ± 0.3 \\ 
 NTK, MLP  & 48.9 ± 1.3 & 47.7 ± 0.4 & 50.9 ± 0.5  & 50.6 ± 0.2 & 51.8 ± 0.2 & 51.6 ± 0.3  \\
 NTK, Linear   & 47.9 ± 1.2 & 47.6 ± 1.2 & 51.4 ± 0.4 & 48.7 ± 0.7 & 52.1 ± 0.3 & 49.3 ± 0.7 \\\hline
 Random, Gaussian & 38.9 ± 0.4 & 49.6 ± 0.9 & 36.2 ± 1.4 & 48.9 ± 0.5 & 29.6 ± 2.3 & 48.7 ± 0.5 \\
 Random, Rademacher & 38.8 ± 0.1 & 49.5 ± 0.4 & 35.3 ± 0.6 & 49.0 ± 0.7 & 30.2 ± 2.1 & 49.0 ± 0.8 \\ \hline 
\end{tabular}}
\end{table}

\subsection{Consistency across model size and datasets}\label{consistency}

\paragraph{Wide ResNets further show the curse of dimensionality.} We study the effect of parameter sizes on our findings by applying different Target and Guess pairs on Wide ResNets. We provide in Tab. \ref{tab:wideresnet-global} the accuracies we obtain for different guesses for a Global Target with a Wide ResNet-18 for different width factors $k$ (for $k=0.5$, $2$, and $4$, with $k=1$ being the standard ResNet-18 we studied earlier), by scaling the number of features according to $k$ (for both the model and the auxiliary network). We observe that Local Guess accuracy improves with width. However, Random Guess accuracy decreases, confirming that Forward Gradient estimation with random directions deteriorates as the size of the gradient increases due to the curse of dimensionality. This in turn suggests pursuing good direction guesses. We also provide in Tab. \ref{tab:wideresnet-local} in App. \ref{localtargetapp} results for a Local Target.

\paragraph{Our findings extend to other datasets.}\label{add datasets} We also showcase results for other image classification datasets with varying difficulty.
We consider the following datasets: Fashion-MNIST, a more complex replacement for MNIST, CIFAR-100, the same dataset as CIFAR-10 with $100$ labels, and Imagenette, an easier subset of Imagenet with only $10$ labels. We summarize the accuracies for a Global Target in Tab. \ref{tab:imagenette-global} and a Local Target in Tab. \ref{tab:imagenette-local} in App. \ref{localtargetapp}  for a ResNet-18 divided into 8 and 16 blocks. We notice the same trends discussed in this study for CIFAR-10, thus confirming our findings.

\section{Conclusion}
We have extensively studied various Forward Gradient training variants for a ResNet-18 architecture divided into 8 and 16 blocks on the standard object recognition tasks CIFAR-10 and ImageNet32. In particular, we introduce the use of gradients from locally supervised auxiliary losses as a better candidate direction for the Target than random noise. We varied Gradient Targets, Guesses, auxiliary networks, and feedback insertion points. The Gradient Guesses selected were either Random isotropic directions, NTK gradients, or local auxiliary net gradients. We confirmed our findings for other image datasets and varying model sizes.

We determined several unambiguous trends. Firstly, Gradient Guesses obtained from Local Guesses exceeded the performance of Random Guesses.
Secondly, our study confirms that estimating consistently the Global Target should be the main focus of Forward Gradient algorithms.
Thirdly, we conclude that the limits of our method are due to the limited alignment between the local loss gradients and end-to-end gradients. 
Still, we found that Local Gradient Guesses reduces the gap with end-to-end training for Forward Gradient and that subsequent works need to improve alignment between Guess and Target to reach end-to-end accuracy.

\section*{Acknowledgements}

This work was supported by Project ANR-21-CE23-0030 ADONIS, EMERG-ADONIS from Alliance SU, and Sorbonne Center for Artificial Intelligence (SCAI) of Sorbonne University (IDEX SUPER 11-IDEX-0004). This work was granted access to the AI resources of IDRIS under the allocations 2022-AD011013095 and 2022-AD011013769 made by GENCI. We acknowledge funding and support from NSERC Discovery
Grant RGPIN-2021-04104 and resources from Compute Canada and
Calcul Quebec. 

\bibliographystyle{icml2023}
\bibliography{biblio}

\clearpage
\appendix
\section{Additional details}
\label{appendix}
\subsection{ResNet-18 Design}

We follow the standard ResNet-18 implementation when decoupling with $8$ blocks, and remove all skip-connections in the $16$ blocks case as explained. For all datasets (except Imagenette, see Sec. \ref{localtargetapp}), we use the standard modification of the first layers of the ResNet to accommodate the smaller image size of our datasets: The first layers are replaced by a convolutional layer with kernel size $3$ and stride $1$ and no bias, a BatchNorm layer with affine output, and a ReLU layer (note that we do not have a Max Pooling layer).

\subsection{Auxiliary Net Design}

The auxiliary nets used to train our model in the experiments are designed to keep the ratio of FLOPS between the auxiliary net and the main module under 10\%. We investigate three architectures: a convolutional neural network (CNN), a multi-layer perceptron (MLP), and a linear classifier. We evaluate each architecture by using it as a local loss to train a ResNet-18 split in 8 blocks with the DGL algorithm. The optimization is carried out by stochastic gradient descent with a momentum of 0.9, a weight decay of $5 \times 10^{-4}$, and mini-batch size 256. The initial learning rate is set to 0.1 and is decayed by a factor of 0.2 every 30 epochs. \\

The linear classifier applies batch normalization followed by an adaptive average pooling which yields an output with spatial resolution $2 \times 2$. This output is then flattened into a 1-dimensional tensor, which is then projected onto the classification space. The MLP architecture first employs an adaptive average pooling yielding an output with spatial resolution $2 \times 2$. This output is then flattened and propagated through a number $n_{\text{depth}}$ of the fully-connected layer using batch normalization and ReLU nonlinearity. Finally, the output is then projected onto the classification space. The first fully-connected layer outputs a vector with $h_{\text{chan}}$ dimension, which parameterizes the width of the auxiliary net, and this number of channels is kept fixed until the projection onto the classification space. The number $n_{\text{depth}}$ of fully-connected layers instead parameterizes the depth of the MLP auxiliary net. Similarly, CNN first employs a 1x1 convolution to obtain a spatial output with $h_{\text{chan}}$, where $h_{\text{chan}}$ essentially parameterizes the width of the architecture. It is followed by a number $n_{\text{depth}}$ of convolutional layers with kernel $3 \times 3$ and stride equal to 2. Using strided convolution helps reduce the computational footprint while maintaining similar accuracies.\\

We tested $n_{\text{depth}}$ between 1 to 8 to test the effect of the depth. We tested $h_{\text{chan}}$ = 64, 128, 256, 512, 1024, 2048, 4096 for the MLP, and $h_{\text{chan}}$ = 4, 8, 16, 32, 64 for CNN. The ResNet-18 was trained on CIFAR-10 using standard data augmentation with the DGL training procedure for 90 epochs. Among the configurations complying with the 10\% FLOPS ratio constraint, we kept the parameters yielding the best accuracy on CIFAR-10. This yields $h_{\text{chan}} = 1024$ and 32 for the MLP and CNN respectively, and $n_{\text{depth}} = 3$ for both architectures.

\subsection{Implementation details}
\paragraph{Practical implementation} In practice, we compute the projection between Guess and Target in our implementation by computing two backward passes, with both losses. For activity perturbation, a backward hook is used to first log the Gradient Target, and then to log the Gradient Guess and compute the projection. We then let the new estimated activation gradient backpropagate through the model. For weight perturbation, we store the (batch-wise) weight gradients after both target and guess backpropagation, and compute weight-wise the projection. 

Of course, both implementations are implementable using forward mode automatic differentiation with modern deep learning libraries. Similarly, we note that for the experiment of Tab. \ref{tab-span}, where we project the Target on a batch of Guesses, this projection is still possible by using $k$ Jacobian-vector products by projecting the Global Target on the $k$ principal components of the batch of Local Guesses. 

\paragraph{Data augmentation} The data augmentation procedure we use is a standard Random Crop with padding $4$ and a Random Horizontal Flip (with probability $0.5$), plus a normalization step. Fashion-MNIST do not use any data augmentation.

\paragraph{Learning rates chosen} In practice, the learning rate (lr) chosen for CIFAR-10 is $0.005$ for Random Guesses (except for the activity-perturbed with a Global Target which has a lr of $0.01$), as well as the 16-blocks Local Linear Guess (activity-perturbed). NTK Guesses uses a lr of $0.01$ in the 16 blocks case (as for the Wide ResNet). With all other Gradient Guesses, the lr is $0.05$.

For the additional CIFAR-100, Fashion-MNIST and Imagenette datasets, end-to-end learning, local learning and local guesses use a lr of $0.05$. CIFAR-100 uses the same lr as CIFAR-10 for the NTK guesses. Fashion-MNIST uses a lr of $0.01$ for activity-perturbed NTK guesses and $0.05$ for the weight-perturbed ones. Imagenette uses a lr of $0.05$ for all NTK guesses. The Random Guesses have a lr of $0.005$.

ImageNet32 proved trickier to optimize. The learning rate and scheduler steps size chosen for the reported accuracy are: For the End-to-End training, a lr of $0.05$ with step size $20$. For the 16 blocks model, local guess activity perturbed have a lr of $0.01$ with step size $30$ and weight perturbed a lr of $0.1$ with step size $20$. The NTK guess have a lr of $0.005$ with step size $30$ (except the CNN auxiliary, with lr $0.01$). Random guesses have a step size of $30$ with a lr of $0.001$ for the activity perturbed and $0.005$ for the weight perturbed. For the 8 blocks model, local guesses have a lr of $0.05$ for the CNN auxiliary, $0.01$ for the linear auxiliary and the activity-perturbed MLP auxiliary, and $0.1$ for the weight-perturbed MLP auxiliary. Activity-perturbed CNN and Linear auxiliary and the weight-perturbed MLP auxiliary have a step size of $20$, and the other $30$. NTK guesses have a step size of $30$ and lr of $0.001$ for the CNN auxiliary, and the MLP and Linear activity-perturbed auxiliaries. The others have a lr of $0.005$.

\section{Additional results}

\subsection{Local Target results} \label{localtargetapp}

We provide in this subsection the Table results discussed in Sec. \ref{consistency}.

\paragraph{Additional datasets} We report in Tab.  \ref{tab:imagenette-local} the results of different guesses with a Local Target for the additional datasets Fashion-MNIST, CIFAR-100, and Imagenette. The results are consistent with our findings on CIFAR-10 and ImageNet32.
We revert the first layers of the ResNet-18 to its original design with Imagenette since its image size is bigger than our other datasets.

\paragraph{Wide ResNet} We report in Tab. \ref{tab:wideresnet-local} the results of different guesses with a Local Target for a Wide ResNet-18 with width factors $k=0.5, 2$ and $4$. We observe similar tendencies to the ones observed for the Global Target.

\subsection{Additional local losses and guesses}\label{fixntk}

\paragraph{Predsim}
To propose a slightly different supervised loss, we refer to the setup of \citet{nokland2019training}, and more particularly the `predsim' local loss for comparison with our more simple cross-entropy local losses. We directly adapt the architecture of the predsim local auxiliary net and loss to our framework, as a Target or Guess. Using predsim for local learning corresponds to the Local Error Signal framework. We report the accuracies using the predsim local loss in Tab. \ref{tab:predsim}. We find predsim to be competitive with the deeper local losses we used (for 8 blocks), despite the predictive auxiliary of predsim being linear. We also note that despite the Global Target loss being different from the Local Loss in that case (minimizing a supplementary similarity matching term), the Local Guess method does not seem affected.
\paragraph{Fixed NTK}

One can note that the dynamics of cosine similarity when training with Local Guess and Global Target (see Fig. \ref{fig cossim activ} and \ref{fig cossim weight}) are similar to the alignment behavior of Direct Feedback Alignment (DFA) which can be observed in \cite{refinetti2021align}. We can thus propose an alternative to our random NTK Guess that is even further inspired by DFA to produce similar gradient feedback. 
Rather than reinitializing the local loss at each batch randomly, we keep a single fixed random local loss throughout training. In this case, the linear Fixed NTK local guess can be computed with no backpropagation and seen as close to DFA.
We use the Fixed NTK local gradient as a type of Local Guess for Forward Gradients to estimate the Global Target. We report in Tab. \ref{tab:fixedntk} the test accuracy obtained for a Resnet-18 on CIFAR-10, for the three types of auxiliary networks. Despite the much more limited Gradient Guess space compared to the random NTK, results are competitive between the two methods. Local learning results also show strong accuracy despite the auxiliary network being fixed compared to the local learning we proposed in Tab. \ref{tablocal}.

\begin{table*}[!ht]
    \centering
    \caption{Test accuracy of a ResNet-18 using a Local Target, split into 8 or 16 local-loss blocks, on Fashion-MNIST, CIFAR-100 and Imagenette datasets for both activity and weight perturbations. We report the mean and standard deviation over 4 runs.}
    \label{tab:imagenette-local}
\resizebox{2\columnwidth}{!}{
\begin{tabular}{|l|l|c|c|c|c|c|c|c|c|c|c|c|c|}
\hline
Dataset & Model  & \multicolumn{6}{c|}{ 16 blocks}  & \multicolumn{6}{c|}{ 8 blocks} \\
\hline
& Local auxiliary   & \multicolumn{2}{c|}{ CNN} & \multicolumn{2}{c|}{MLP} & \multicolumn{2}{c|}{Linear}  & \multicolumn{2}{c|}{ CNN} & \multicolumn{2}{c|}{MLP} & \multicolumn{2}{c|}{Linear} \\\hline
& Gradient Guess               & Activity & Weight  & Activity & Weight   & Activity & Weight  & Activity & Weight   & Activity & Weight   & Activity & Weight  \\ \hline
Fashion-MNIST & Local learning     & \multicolumn{2}{c|}{93.6 $\pm$ 0.1} & \multicolumn{2}{c|}{94.1 $\pm$ 0.2} & \multicolumn{2}{c|}{93.4 $\pm$ 0.1}  & \multicolumn{2}{c|}{94.1 $\pm$ 0.2} & \multicolumn{2}{c|}{94.4 $\pm$ 0.2} & \multicolumn{2}{c|}{93.8 $\pm$ 0.2} \\\hline
& Gaussian   Guess     &  58.3 $\pm$ 0.5 & 85.4 $\pm$ 0.6  & 68.1 $\pm$ 2.6   & 86.1 $\pm$ 0.7 & 67.6 $\pm$ 1.8 & 88.3 $\pm$ 4.8  & 85.2 $\pm$ 1.5 & 89.0 $\pm$ 0.4   & 87.1 $\pm$ 0.7   & 88.9 $\pm$ 0.2  & 82.8 $\pm$ 1.4  & 90.3 $\pm$ 0.1 \\
& Radem.  Guess   & 61.6 $\pm$ 5.5  & 85.3 $\pm$ 0.5  & 66.7 $\pm$ 1.9 & 86.2 $\pm$ 0.5  & 65.0 $\pm$ 2.3 & 87.9 $\pm$ 0.1 &  81.5 $\pm$ 7.7 & 89.1 $\pm$ 0.1  & 86.4 $\pm$ 0.8 & 89.1 $\pm$ 0.1  & 82.9 $\pm$ 1.5 & 90.3 $\pm$ 0.1 \\ \hline 
CIFAR-100 & Local learning     & \multicolumn{2}{c|}{57.3 $\pm$ 0.3 } & \multicolumn{2}{c|}{64.6 $\pm$ 0.3 } & \multicolumn{2}{c|}{62.5 $\pm$ 0.2}  & \multicolumn{2}{c|}{67.3 $\pm$ 0.2} & \multicolumn{2}{c|}{70.3 $\pm$ 0.5 } & \multicolumn{2}{c|}{65.8 $\pm$ 0.5} \\\hline
& Gaussian Guess       &  3.7 $\pm$ 0.6 & 10.7 $\pm$ 0.4 & 3.3 $\pm$ 0.8 & 11.2 $\pm$ 1.2 &3.9 $\pm$ 0.8 &  14.3 $\pm$ 2.7 & 15.7 $\pm$ 1.2 & 25.3 $\pm$ 0.4 & 6.6 $\pm$ 1.8 & 24.9 $\pm$ 0.7  & 11.8 $\pm$ 2.1 &  26.3 $\pm$ 0.4 \\
& Radem. Guess    & 4.4 $\pm$ 0.3 & 7.2 $\pm$ 1.5  & 3.2 $\pm$ 0.8 &  10.4 $\pm$ 1.1 & 4.2 $\pm$ 0.7 &  12.1 $\pm$ 0.3 & 14.4 $\pm$ 2.1 & 24.0 $\pm$  0.9 & 10.7 $\pm$ 2.0 & 23.1 $\pm$ 0.5 & 4.2 $\pm$ 0.7 &  24.4  $\pm$ 0.3 \\ \hline 
Imagenette &  Local learning       & \multicolumn{2}{c|}{84.6 $\pm$ 0.3 } & \multicolumn{2}{c|}{82.4 $\pm$ 0.2  } & \multicolumn{2}{c|}{83.2 $\pm$ 0.6 }  & \multicolumn{2}{c|}{88.6 $\pm$ 0.4 } & \multicolumn{2}{c|}{86.0 $\pm$ 0.3 } & \multicolumn{2}{c|}{86.2 $\pm$ 0.1} \\  \hline 
& Gaussian   Guess         &  21.6 $\pm$ 2.8   & 36.7 $\pm$ 1.5   &    21.0 $\pm$ 2.9 & 36.8 $\pm$ 1.5 & 25.8 $\pm$ 2.3  & 39.0 $\pm$ 0.8  &39.1 $\pm$ 4.4 & 55.6 $\pm$ 0.8 & 44.7 $\pm$ 1.9 & 55.9 $\pm$ 0.4 & 36.5 $\pm$ 4.7 &  58.2 $\pm$ 1.2 \\
& Radem.  Guess    &  22.8 $\pm$ 3.2 & 36.3 $\pm$ 1.3 & 20.3 $\pm$ 2.7 & 37.6 $\pm$ 2.5 & 24.9 $\pm$ 1.8 & 40.1 $\pm$ 0.8 & 38.5 $\pm$ 1.9 & 55.9 $\pm$ 0.8 & 42.3 $\pm$ 0.8 & 56.1 $\pm$ 1.2 & 37.4 $\pm$ 3.2 & 59.0 $\pm$ 1.1 \\ \hline 
\end{tabular}
}
\end{table*}

\begin{table*}[!ht]
    \centering
    \caption{Test accuracy of a Wide ResNet-18 using a Local Target, split into 16 local-loss blocks, on CIFAR-10 for both activity and weight perturbations for different width factors, i.e. k=0.5, 2 and 4. Table \ref{tablocal} refers to the case where $k=1$. We report the mean and standard deviation over 4 runs.} 
    \label{tab:wideresnet-local}
\begin{tabular}{|l|l|c|c|c|c|c|c|}
\hline
Width & Local auxiliary   & \multicolumn{2}{c|}{ CNN} & \multicolumn{2}{c|}{MLP} & \multicolumn{2}{c|}{Linear} \\\hline
& Gradient Guess               & Activity & Weight & Activity & Weight  & Activity & Weight\\ \hline
$0.5$ & Local learning     & \multicolumn{2}{c|}{ 81.1 $\pm$ 0.4 	} & \multicolumn{2}{c|}{ 86.9 $\pm$ 0.4 } & \multicolumn{2}{c|}{ 83.5 $\pm$ 0.4}   \\\hline
& Gaussian   Guess     & 21.0 $\pm$ 2.2   & 34.1 $\pm$ 2.3   &  23.2 $\pm$ 4.0  & 37.7 $\pm$ 0.9   & 13.1 $\pm$ 6.3  & 44.1 $\pm$ 1.0    \\
& Rademacher  Guess   & 21.8 $\pm$ 3.5    &  34.4 $\pm$ 1.8   & 24.0 $\pm$ 1.2   &  38.3 $\pm$ 0.4   &  13.0 $\pm$ 6.4 & 43.0 $\pm$ 0.9  \\ \hline 
$2$ & Local Learning     & \multicolumn{2}{c|}{ 91.1  $\pm$  0.1	} & \multicolumn{2}{c|}{ 	90.8 $\pm$  0.1	} & \multicolumn{2}{c|}{88.5 $\pm$  0.1} \\\hline
& Gaussian Guess       &  17.9 $\pm$ 1.0   & 40.8 $\pm$ 1.0   &    20.5 $\pm$ 2.1 & 42.0 $\pm$ 0.6 & 24.9 $\pm$ 7.8  & 44.3 $\pm$ 1.1  \\
& Rademacher Guess    &  17.9 $\pm$ 2.6  &  41.0 $\pm$ 0.4   &  20.3 $\pm$ 1.4  & 41.7 $\pm$ 0.4  & 23.1 $\pm$ 2.3 & 44.0 $\pm$ 2.6   \\ \hline 
$4$ &  Local learning       & \multicolumn{2}{c|}{93.1 $\pm$  0.1	} & \multicolumn{2}{c|}{ 91.7 $\pm$  0.3	} & \multicolumn{2}{c|}{	 89.4  $\pm$  0.1 }   \\  \hline 
& Gaussian   Guess          & 18.7 $\pm$ 2.8   &  41.8 $\pm$ 0.5  & 18.6 $\pm$ 3.4 & 43.1 $\pm$ 0.0    &  23.2 $\pm$ 2.1 & 44.7 $\pm$ 0.6  \\
& Rademacher Guess    & 17.3 $\pm$ 3.0 &  42.3 $\pm$ 0.4   & 18.6 $\pm$ 2.4 & 42.8 $\pm$ 0.5   &  20.7 $\pm$ 2.9  & 44.8 $\pm$ 0.5  \\ \hline 
\end{tabular}
\end{table*}

\begin{table*}[!ht]
    \centering
    \caption{Test accuracy of a ResNet-18 split into 8 or 16 local-loss blocks, using a Fixed NTK (Fixed randomly parameterized local loss) as a Gradient Guess for the Global Target. }
\label{tab:fixedntk}
\begin{tabular}{|l|c|c|c|c|}
\hline Model &  \multicolumn{2}{c|}{8 blocks} &  \multicolumn{2}{c|}{16 blocks}    \\ \hline
Gradient Guesses & Activity  & Weight & Activity  & Weight  \\ \hline
Fixed NTK, CNN                    & 51.2 & 57.2 & 27.5 & 45.1  \\ 
Fixed NTK, MLP                    & 52.3 & 64.6 & 40.2 & 45.5  \\
Fixed NTK, Linear                      & 57.7 & 76.4 & 36.9 & 60.5\\\hline
\end{tabular}

\end{table*}

\begin{table*}[!ht]
    \centering
    \caption{Test accuracy of a ResNet-18 using a predsim auxiliary loss, on CIFAR-10 for both  activity and weight perturbations. The Local learning case corresponds to the Local Error Signals framework.}
\label{tab:predsim}
\begin{tabular}{|l|c|c|c|c|}
\hline Model &  \multicolumn{2}{c|}{8 blocks} &  \multicolumn{2}{c|}{16 blocks}    \\ \hline
Global Target & Activity & Weight & Activity & Weight \\ \hline
  Local Guess                     & 84.0 & 89.8 & 64.8 & 85.3 \\
 NTK Guess                     & 31.8 & 65.7 & 11.5 & 45.5 \\\hline
Local Target & Activity  & Weight & Activity  & Weight \\ \hline
Local learning  &\multicolumn{2}{c|}{89.2} &\multicolumn{2}{c|}{86.8} \\\hline
 Gaussian Guess   & 56.2 & 37.9 & 37.3 & 30.8\\
Rademacher Guess        & 55.9 & 39.4 & 37.2& 30.2\\ \hline 
\end{tabular}

\end{table*}

\subsection{Figures for a ResNet-18 split in 16 blocks}

We also provide additional Figures equivalent to the Figures in the main paper but for a ResNet-18 trained in 16 blocks and without skip connection.

\begin{figure}[ht]
          \centering
          \includegraphics[width=.95\linewidth]{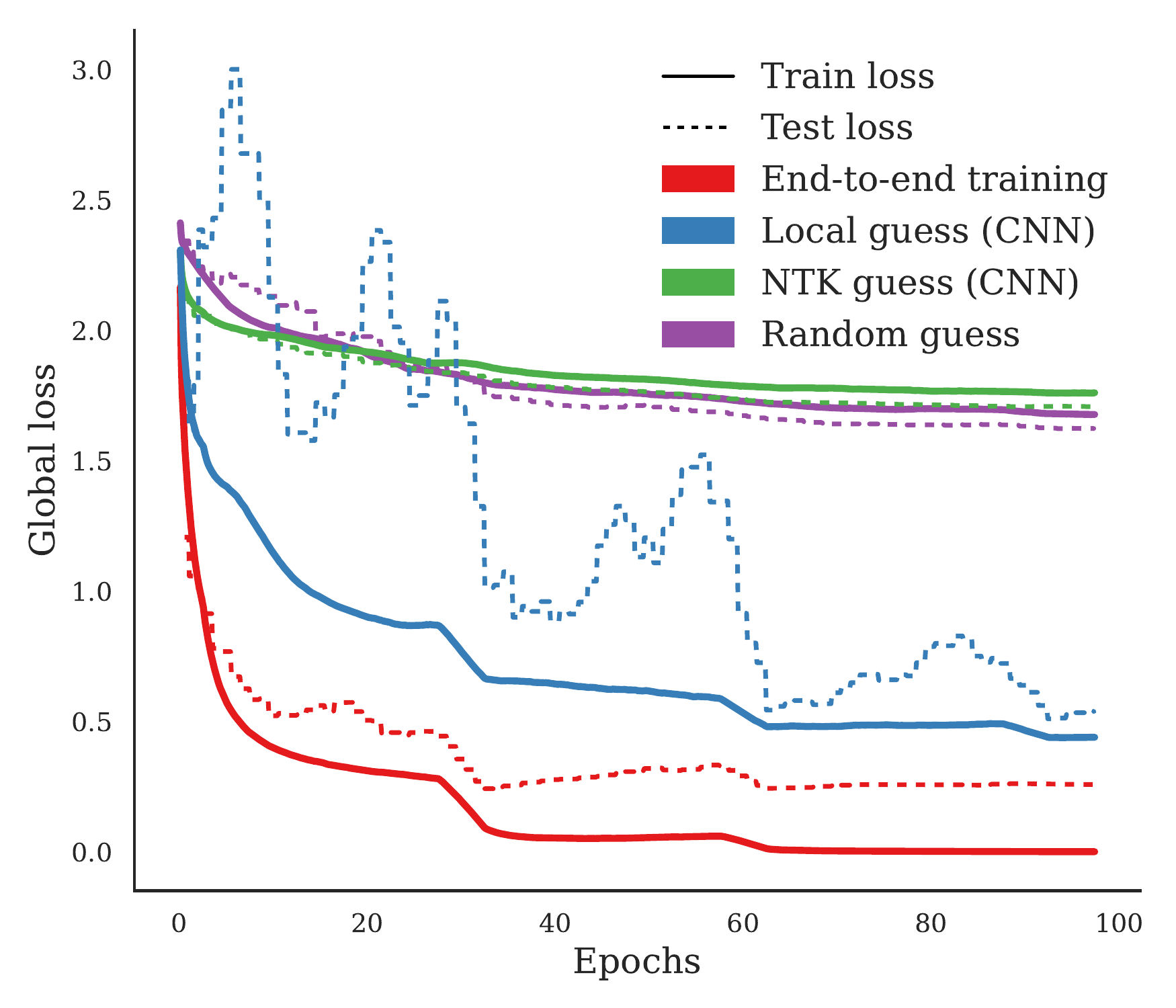}
          \caption{Comparison of train and test losses for end-to-end training (red), and Forward Gradient with Global Target and Local Guess (blue), NTK Guess (green) and Random Guess (purple), on CIFAR-10 with a ResNet-18 split in 16 blocks. Local Guess and NTK Guess are derived from a CNN auxiliary.}
\end{figure}

        \begin{figure*}[ht]
          \centering
          \includegraphics[width=.5\linewidth]{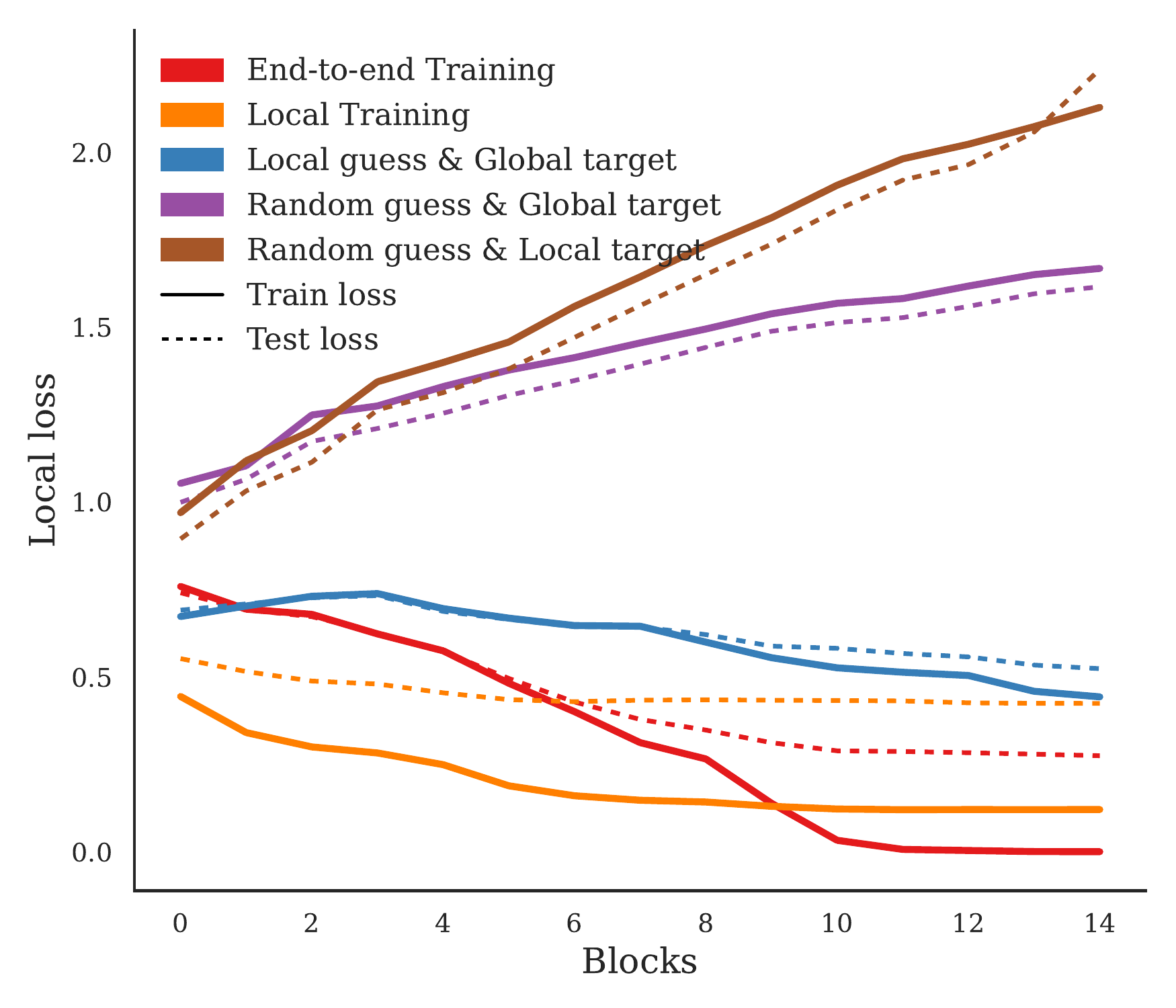}%
          
          \caption{Local train losses at the end of training at each block for a ResNet-18 split into 16 blocks, with CNN auxiliary, for different training algorithms. In the Gaussian and End-to-End cases, the auxiliary training is detached from the main module training and is only for logging purposes.}
\end{figure*}

 \begin{figure*}[ht]
          \centering
          \includegraphics[width=.8\linewidth]{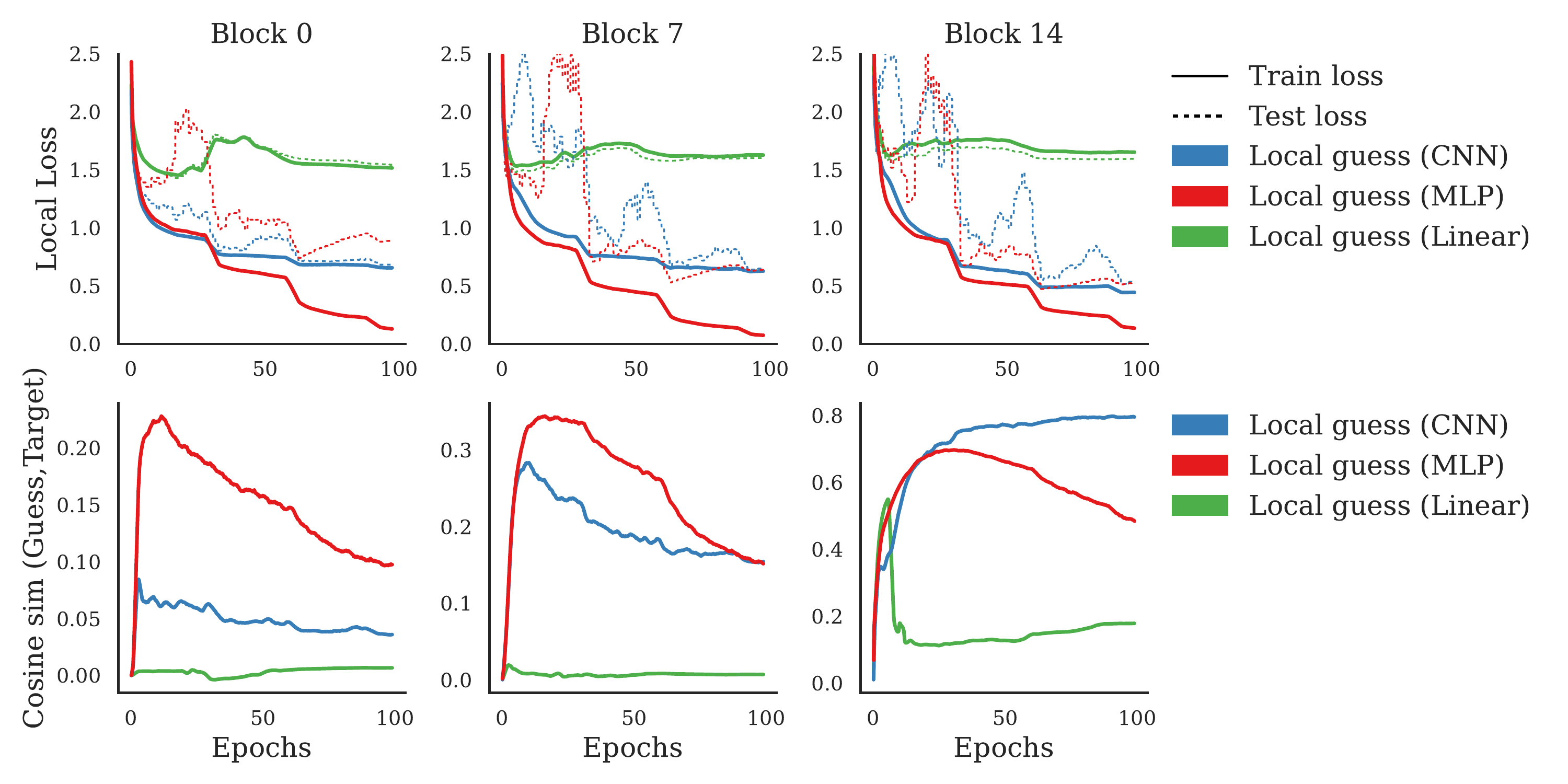}%
          
          \caption{Train and test local losses (top row) and mean cosine similarity between a Local Guess and Global Target in the activation space (bottom row), for blocks 0, 7, and 14 (left, middle, and right columns) during training. The model is a ResNet-18 divided into 16 blocks trained on CIFAR-10.}
        \end{figure*}

        \begin{figure*}[ht]
          \centering
          \includegraphics[width=.8\linewidth]{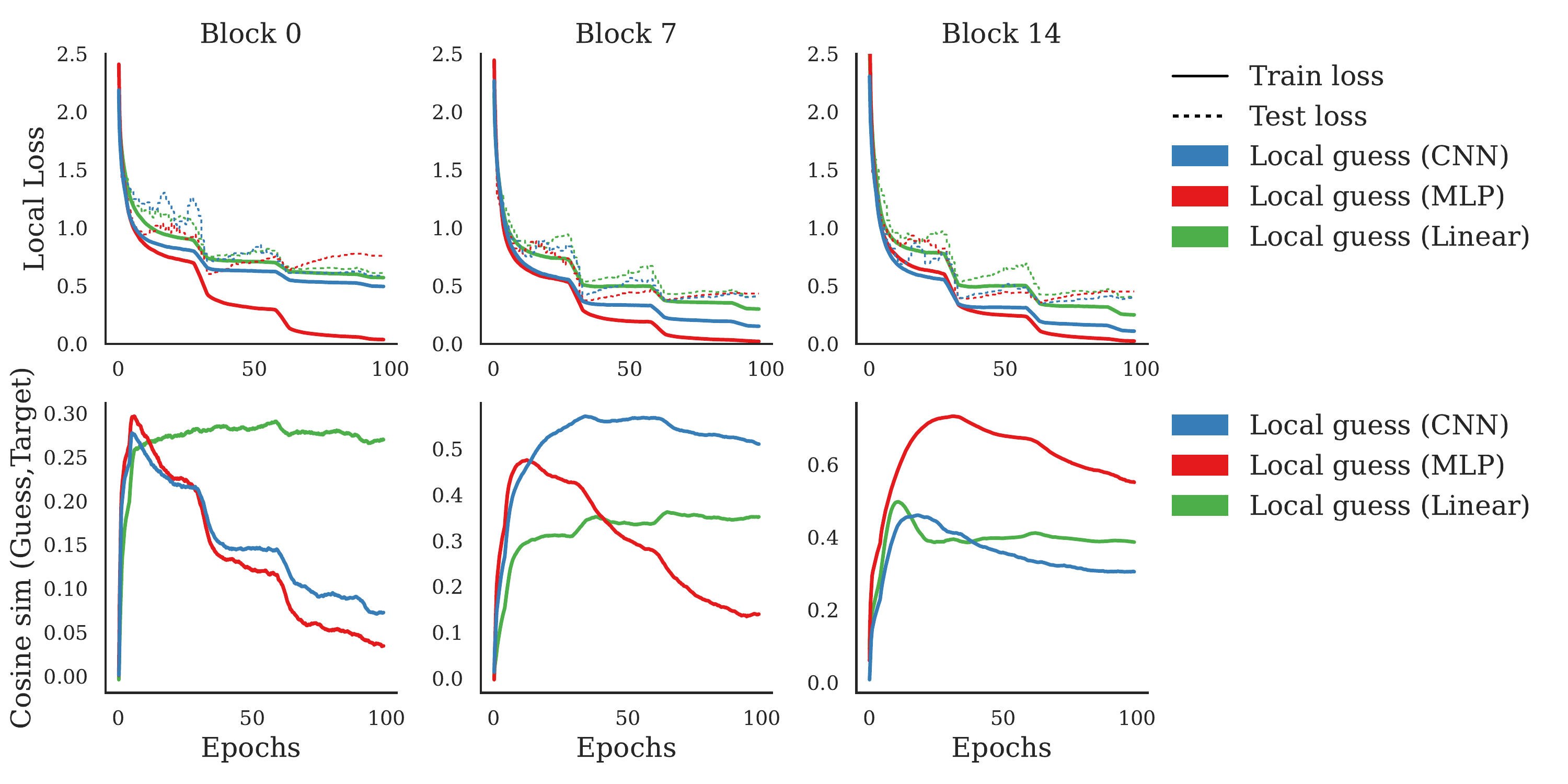}
          \caption{Train and test local losses (top row) and mean cosine similarity between a local guess and global target weight gradients (bottom row, averaged over the parameters), for blocks 0, 7, and 14 (left, middle, and right columns) during training. The generalization gap is more consistent during training than with activation gradients. The cosine similarity is also consistently higher than activation gradients but falls more drastically during training. The model is a ResNet-18 divided into 16 blocks trained on CIFAR-10.}
        \end{figure*}

\end{document}